\documentclass[runningheads]{llncs}

\usepackage{eccv}

\usepackage{eccvabbrv}
\usepackage{wrapfig}
\usepackage{multirow}
\usepackage{soul}
\usepackage{graphicx}
\usepackage{booktabs}

\usepackage[accsupp]{axessibility}  

\usepackage{colortbl}
\usepackage{makecell}
\usepackage{multirow}
\usepackage{booktabs}
\usepackage{hhline}
\usepackage{algorithm}
\usepackage{algorithmic}

\definecolor{LGray}{gray}{0.88}
\sethlcolor{LGray}

\definecolor{mygray}{gray}{.9}
\definecolor{mygreen}{rgb}{0, 0.6, 0}
\definecolor{mygreen}{rgb}{0.0, 0.5, 0.0}
\definecolor{myteal}{rgb}{0.16, 0.47, 0.56}
\definecolor{mypink}{rgb}{0.81, 0.25, 0.44}
\definecolor{stepcolor}{RGB}{70,130,180}
\definecolor{MySoftBlue}{HTML}{002E90}
\definecolor{MySoftPink}{HTML}{660000}
\definecolor{ao(english)}{rgb}{0.0, 0.5, 0.0}
\definecolor{Gray}{gray}{0.8}
\definecolor{RedOrange}{RGB}{255,69,0}
\definecolor{mypink}{RGB}{255, 105, 180}
\definecolor{blue2}{HTML}{0066CC}
\definecolor{orange2}{HTML}{CC6600}

\def\eg{{\em e.g. }}
\def\ie{{\em i.e. }}
\def\etc{{\em etc. }}
\def\etal{{\em et al.}}

\def\Er{{\color{red} ${\mathcal{A}_\mathcal{E}}$ }}
\def\MER{{\color{red} {\mathcal{A}_\mathcal{E}} }}
\def\EER{{\color{red} {\mathcal{E}} }}
\def\SUPP{{\em Supplementary}}

\newcommand{\COMM}[1]{{\color{MySoftBlue} #1 }}
\newcommand{\COM}[1]{{\color{purple} \textit{#1} }}
\newcommand{\OBJECT}[1]{\textbf{Objective:} #1}

\newcommand{\TODO}[1]{\textbf{\color{red}[TODO: #1]}}
\renewcommand{\TODO}[1]{}

\def\CP{{\ttfamily{CoSPlan}}} 
\def\CPP{\CP { }} 

\DeclareMathOperator{\ER}{\mathbb{E}}%
\newcommand{\HB}[1]
{{#1}}

\newcommand{\RC}[1]
{{\cellcolor{red!15}#1}}
\newcommand{\WR}[1]
{{\cellcolor{red!30}#1}}
\newcommand{\BR}[1]
{{\cellcolor{green!20}#1}}

\newcommand{\RCTEXT}[1]{{{\sethlcolor{red!15} \hl{#1}}}}
\newcommand{\WRTEXT}[1]{{{\sethlcolor{red!35} \hl{#1}}}}

\newcommand{\figno}[1]{{\color{blue}#1}}
\newcommand{\insight}[1]{\hl{#1}}

\newcommand{\RED}[1]{{\color{red}#1}}
\newcommand{\BLUE}[1]{{\color{blue2}#1}}
\newcommand{\GREEN}[1]{{\color{ao(english)}#1}}

\usepackage{hyperref}

\usepackage{orcidlink}

\newcommand\blfootnote[1]{%
  \begingroup
  \renewcommand\thefootnote{}\footnote{#1}%
  \addtocounter{footnote}{-1}%
  \endgroup
}

\begin{document}

\title{\CP: \underline{Co}rrective \underline{S}equence \underline{Plan}ning via Scene Graph Incremental Updates}

\titlerunning{\CPP}

\author{Shresth Grover$^\dagger$\and
Priyank Pathak \and
Akash Kumar\and 
Yogesh S Rawat
}
\authorrunning{Grover et al.}

\institute{
UCF Institute of Artificial Intelligence, University of Central Florida (UCF)\\[1ex]
\email{shgrover@ucsd.edu}
\hspace{2pt}
\email{\{priyank, akash.kumar, yogesh\}@ucf.edu}\\[1ex]
\textbf{Project Page}: \url{https://shroglck.github.io/cos_plan/} \\
\textbf{Dataset}: \url{https://huggingface.co/datasets/shrg7/COSPLAN} 
\blfootnote{$^\dagger$work done as intern}
}

\maketitle

\begin{abstract}
Vision Language Models (VLMs) have shown promising planning capabilities, yet 
their success remains confined to 
the text domain, 
leaving visual decision-making relatively underexplored.
Addressing this gap, we introduce \textbf{\underline{\textbf{Co}}rrective \underline{\textbf{S}}equence \underline{\textbf{Plan}}ning  (\textbf{\CP})} benchmark, where VLMs must plan a sequence of visual actions from an initial scene to a target scene.
\CPP evaluates models on their ability to imagine and execute a coherent set of visual steps required to reach the goal (\textbf{Step Completion}).
To prevent any shortcuts that simply describe the final scene, we introduce an erroneous action in decision making, which must be \textit{detected} (\textbf{Error Detection}) and \textit{corrected} to reach the goal, enabling a deeper understanding of the task. 
\CPP spans across 4 tasks: maze navigation, block re-arrangement, image reconstruction, and object re-organization.
Despite using advanced reasoning strategies such as \textit{Chain-of-Thought} and \textit{Scene Graphs}, VLMs struggle on \CP, while still showing promising performance in the text domain. 
Addressing this, we propose \textbf{\underline{\textbf{S}}cene \underline{\textbf{G}}raph \underline{\textbf{I}}ncremental updates (\textbf{SGI})}, a novel training-free method to transform images into \textit{`textual'} scene graphs, enabling step-by-step reasoning through iterative scene graph refinement. 
SGI yields an average of  $\simeq4.4\%\uparrow$ on \CPP w/ generalization on PlanBench and VQA.
\textit{Link for solving puzzles on the project page.}
\end{abstract}

\section{Introduction}

Humans are uniquely equipped to handle visual-language information to solve complex problems by predicting future steps under physical constraints.
In AI, Large-scale Vision Language Models (VLMs)~\cite{gpt4, janus} replicate a similar mechanism in the textual domain~\cite{huang2024large, valmeekam2023planning}. 
Using techniques such as Chain-of-Thought (CoT) and deep-thinking paradigms, VLMs can generate a sequence of textual steps enabling reasoning over long sequences of instructions.
This success in text-based decision making raises a critical but \textbf{vastly unexplored question}: \textit{``What happens when this sequential thought process must occur in the visual domain?"} 

\begin{figure*}[!t]
\centering
\begin{subfigure}{.44\textwidth}
\centering
\includegraphics[width=\linewidth]{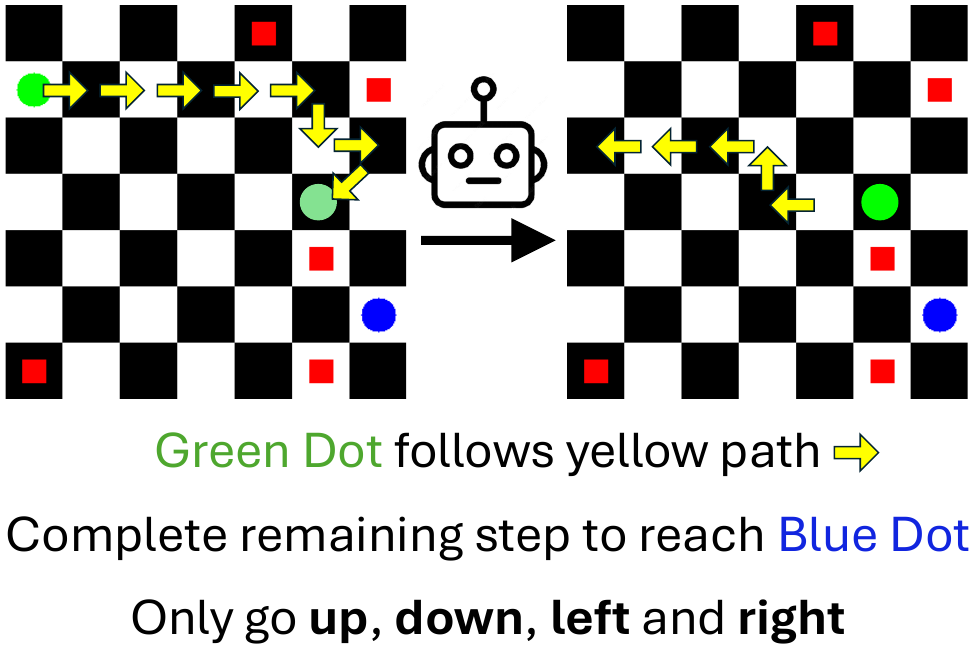}
\end{subfigure}
\hfill
\begin{subfigure}{.54\textwidth}
\centering
\includegraphics[width=\linewidth]{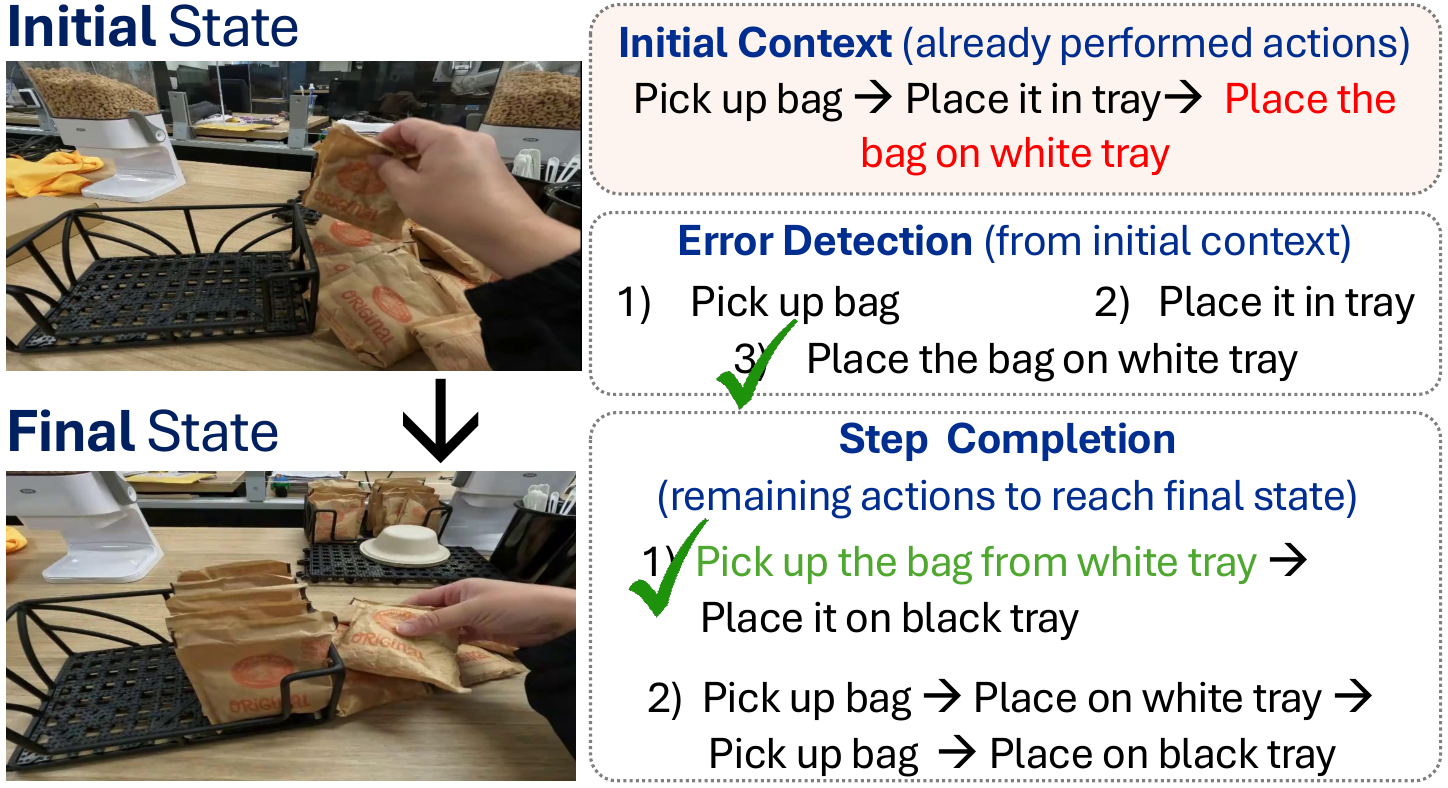}
\end{subfigure}
\caption{\textbf{\CPP}: 
\textit{(Left)} Given the initial \textit{multi-step instructions} (yellow), VLM is asked to complete the remaining steps to reach blue dot. 
Errors include starting at the wrong position (green dot shifted by 1 cell), and choosing the path because it's the first option among all possible paths.
\textit{(Right)} The model sees the initial \& final states, along with an initial context of already performed actions (w/ an \RED{error}). 
VLM identifies the error and picks the set of remaining actions (\GREEN{correcting error}) to reach goal.
}
\label{fig:paper_overview}
\end{figure*}

VLM's strong zero-shot generalization is ideal for complex visual reasoning, a prominent feature in real-world decision-making (\eg robotics, autonomous navigation \textit{etc}).
These models can accurately 
predict (imagine) some future scenes, yet they struggle to sustain coherent multi-step visual reasoning (\figno{\cref{fig:paper_overview} (left)}).
Moreover, explicitly visualizing (generating) future scenes is computationally expensive, making naive simulation impractical for real-world deployment.

Recent benchmarks focus on simple visual puzzles for evaluating logic reasoning~\cite{lu2023mathvista, kamoi2024visonlyqa,wu2024vspassessingdualchallenges}, but lack complex decision making. 
In contrast, decision-making problems like  \textit{sequential planning} require a step-by-step plan to reach a goal~\cite{paul2023sequentialplanninglargepartially, wang2024qimprovingmultistepreasoning, nayak2024mapthor}. However, research on such planning tasks is overwhelmingly confined to the textual domain~\cite{valmeekam2023planning, ramakrishnan2024does}, with limited-to-no exploration in vision~\cite{illusion-of-thinking}.
Bridging this gap, we focus on \textit{vision}-based logical reasoning guided by \textit{textual} instructions for decision making, involving \textit{temporal} sequence of actions.
Our 
\textbf{\underline{\textbf{Co}}rrective \underline{\textbf{S}}equence \underline{\textbf{Plan}}ning}
(\textbf{\CP}) is designed to evaluate VLMs' reasoning abilities to complete tasks through multi-step visual actions. 
Starting from an initial scene and a sequence of actions, the model must understand the context and determine a plan that correctly completes the remaining steps to reach the goal (\textbf{Step Completion}).
This requires not only understanding the scene and its constraints, but also updating the scene after each action.

Unlike ideal settings with perfect instructions~\cite{sener2022assembly101}, \CPP reflects realistic conditions by introducing an \textit{intentional erroneous step} in the action sequence.
Models are then judged on their 
ability to perform \textbf{Error Detection}, 
requiring models to identify the incorrect step, correct it, and reach the goal.
This not only \textit{ensures a genuine understanding of the task}, where models must \textit{scrutinize  each and every action}, 
but also \textbf{prevents cheating}~\cite{shaib2026learningwronglessonssyntacticdomain,eshuijs2025shortcircuitingshortcutsmechanisticinvestigation}, where models can't simply select an option that matches the final state (bypassing true reasoning).
Errors also reflect practical scenarios where long sequences of actions may introduce errors. 
An example of \CPP is shown in\figno{~\cref{fig:paper_overview} (right)}.

\CPP introduces 4 error-prone planning tasks:
1) \textbf{Maze-E} for navigation,
2) \textbf{Blocks-World-E} for block re-arrangement, 
3) \textbf{Shuffle-E} for image re-construction, and 
4) \textbf{Robo-VQA-E} for real-world re-organization.
We evaluate several VLMs, including \textit{GPT-4o} and GPT-5.1~\cite{gpt4}, \textit{CoG-VLM}~\cite{cogvlm}, \textit{Intern-VLM} variants~\cite{internvl,wang2025internvl35advancingopensourcemultimodal}, \textit{Janus-Pro-7B}~\cite{janus}, \textit{Qwen} variants~\cite{qwen, bai2025qwen3vltechnicalreport}, and Llama3-8B\cite{grattafiori2024llama3herdmodels}.
Their vanilla performance is near 
random guessing, indicating the difficulty of the benchmark.
Hence, two deeper reasoning strategies are applied: Chain-of-Thought (CoT)~\cite{wei2022chain} to encourage step-wise reasoning, and Scene Graph (SG)~\cite{cot_reasoning_vlm2023} to structure object positions and attributes.
This setup highlights the benchmark’s complexity and provides a rigorous test of visual reasoning ability.

While CoT \& SG are known to perform well on error-free planning tasks, we show that they struggle on \CP. 
They compress reasoning from initial to final state into a single step. 
This forces  models to implicitly simulate (\textit{internalize}) long action sequences without explicit intermediate representations.
To address this, we introduce 
\textbf{SGI (\underline{\textbf{S}}cene \underline{\textbf{G}}raph \underline{\textbf{I}}ncremental updates)}, a training-free method 
that represents \textit{images as text-based Scene Graphs} and \textit{refines them step-by-step}, explicitly generating intermediate states.
Thus, SGI i) reduces long horizon reasoning burden, ii) tracks evolving scenes, and iii) detects and corrects errors, making corrective sequence planning significantly more robust.

In summary, we make the following contributions: i) 
\textbf{\CP} (\underline{\textbf{Co}}rrective \underline{\textbf{S}}equence \underline{\textbf{Plan}}ning) is the first multimodal benchmark with temporal sequences of actions in vision + language domain to evaluate VLMs visual reasoning in error-prone scenarios. 
\CPP includes four planning tasks to test the abilities of \textit{Error Detection} and
\textit{Step Completion} to reach a desired goal. 
ii) We evaluate VLMs and reveal weaknesses in error handling, vision-based sequence planning, and understanding. 
iii) We propose \textbf{SGI}, a \underline{\textbf{S}}cene \underline{\textbf{G}}raph \underline{\textbf{I}}ncremental technique, that refines representations of scene for every action, enhancing robustness in \CPP and datasets such as VQA~\cite{wang2024pictureworththousandwords}, and PlanBench~\cite{valmeekam2023planning}. 

\begin{table}[!t]
\centering
\caption{\textbf{Reasoning Benchmarks}: 
Temporal updates the scene with action. The multi-step tasks don't complete after one action. Error is erroneous decision-making. 
}
\renewcommand{\arraystretch}{1}
\setlength\tabcolsep{3pt}
\resizebox{\textwidth}{!}{
\begin{tabular}{c |c c c | c c | c c c}
\specialrule{1.5pt}{0pt}{0pt}
\rowcolor{mygray}
& \multicolumn{3}{c|}{\textbf{Modality}} & \multicolumn{2}{c|}{\textbf{Testing Environment}} &
\multicolumn{2}{c}{\textbf{Task}}
\\
[-.4pt]\hhline{~ --- -- --}

\rowcolor{mygray}
\multirow{-2}{*}{ \textbf{Benchmark}} &

Vision & 
Text &
Temporal & 

Synthetic & 
Real World &  

Multi-step &
Error \\
\hline 
ALFWorld~\cite{shridhar2020alfworld}  & & \checkmark          & \checkmark & \checkmark & & \checkmark \\
PlanBench~\cite{valmeekam2023planning}  & &\checkmark  & \checkmark & \checkmark & &\checkmark &  \\
WebArena~\cite{zhou2024webarenarealisticwebenvironment} & &\checkmark  & \checkmark & \checkmark & & \checkmark & \\
SpatialEval~\cite{wang2024pictureworththousandwords} & \checkmark & \checkmark &  & \checkmark & \checkmark  & &  \\

SURDS~\cite{guo2025surds} &\checkmark & \checkmark & \checkmark &  & \checkmark & &  \\
\hline
\rowcolor{mygray}
\CPP(\textit{Ours})
&\checkmark & \checkmark  & \checkmark & \checkmark & \checkmark & \checkmark & \checkmark \\
\specialrule{1.5pt}{0pt}{0pt}
\end{tabular}
}
\label{tab:llm-planning-benchmarks}
\end{table}

\section{Related work}

\noindent \textbf{Reasoning in VLMs} 
Enhancing Reasoning without fine-tuning~\cite{cheng2024spatialrgpt} is of great interest in VLMs.
Wei~\etal\cite{wei2022chain} introduced the Chain-of-Thought (CoT) via a series of reasoning steps to guide LLMs.
CoT shows significant gain on mathematical reasoning tasks~\cite{sprague2024cot}; however, it \textit{lacks spatial relationships}, making it unfit for visual reasoning.
Chen~\etal\cite{cot_reasoning_vlm2023} proposed using Scene Graphs (SG) as structured representations, to improve VLM’s reasoning abilities on visual grounding tasks like VQA \cite{damodaran2021understandingrolescenegraphs}, image generation \cite{johnson2018imagegenerationscenegraphs},
spatial reasoning~\cite{li2021embodied}, \etc
However, SG tries to hallucinate an entire sequence in one go, facing challenges in erroneous multi-step visual reasoning.
Our \textbf{SGI} approach converts images into textual graphs, which VLMs can then iterate over to generate intermediate scenes like CoT, without relying on external annotations (unlike~\cite{rana2023sayplan}).

\vspace{4pt}
\noindent \textbf{Sequential Planning}
Valmeekam~\etal\cite{valmeekam2023planning} proposed variations of Sequence Planning, including completing steps based on a partial context. 
Most sequential planning datasets~\cite{asai2018photorealisticblocksworlddataset, sener2022assembly101, zhang2024ingvpmllmsplayeasy, robovqa, nagpal2025optimalroboticassemblysequence} assume ideal instructions, which may not hold outside the lab environment. 
Many rely on human / video-based supervision~\cite{crockett2025human, zhao2022p3ivprobabilisticprocedureplanning, bouhsain2023learning}, limiting scalability.
Most benchmarks focus on textual domain~\cite{xiao-etal-2024-flowbench, ramakrishnan2024does, zheng2024naturalplanbenchmarkingllms,asai2022classicalplanningdeeplatent}, with limited exploration in the vision domain~\cite{NEURIPS2023_efb2072a, illusion-of-thinking, chow2025physbench} 
(\figno{\cref{tab:llm-planning-benchmarks}}). 
SpatialEval~\cite{wang2024pictureworththousandwords} only evaluates static images with no temporal change of states.
Recent works have also examined VLMs for planning~\cite{kambhampati2024position, hao2023reasoning, huang2024understanding, zhang2024fltrnn, Rossetti_Tummolo_Gerevini_Putelli_Serina_Chiari_Olivato_2024}.
\CPP is the first benchmark to evaluate VLMs on sequential planning under \textit{vision-language and temporal domain} with error-prone instructions.

\vspace{4pt}
\noindent \textbf{Explainable AI \& Analysis}
Explaining behavior VLMs in the real-world has been quite a well-researched topic~\cite{kazmierczak2025explainability, pathak2025lrfm, shu-etal-2025-large, sim-etal-2025-vlms}. However, agentic models like GPT~\cite{gpt4} and Qwen~\cite{qwen} are fairly recent, and their explanations of decision-making are still in their infancy~\cite{grover2024navigatinghallucinationsreasoningunintentional, luo2024understanding, palikhe2025towards}. In this work, we aim to study how these AI models make reasoning decisions when faced with erroneous step completion tasks.  
We leverage \textit{single-choice MCQ} for this, as it aids controlled analysis of VLM decision-making. This further aids easy analysis of failure cases and serves as motivation for our SGI algorithm, 

\begin{table*}[!t]
\centering
\caption{\textbf{\CPP Dataset Details:}
`Size' are the number of image-text pairs,
`Initial Context' is the \textit{average} number of actions already performed, and `Remaining Steps' is the \textit{average} number of additional steps on top of the initial context, required to reach the goal. The `Source' is where images (or text) are taken from. 
}

\setlength\tabcolsep{2pt}
\resizebox{\textwidth}{!}{
\begin{tabular}{l lcccccc}
\specialrule{1.5pt}{0pt}{0pt}
\rowcolor{mygray}  
\textbf{Dataset}  & 
\textbf{Task} & \textbf{Type} & 
\textbf{\makecell{Size}}&
\textbf{\makecell{Initial \\ Context}} & 
\textbf{\makecell{Remaining \\ Steps}} & \textbf{Source }  \\ 
\hline 
Maze-E & Navigation & Path Planning &  5000 & 2.0 & 4.6 & Synthetic  \\ 
Blocks-World-E & Re-arrange & Blocks & 5000 & 2.0 & 3.8 & Synthetic  \\ 
Shuffle-E & Re-construct & Puzzle & 1000 & 3.7 & 7.1 & ImageNet~\cite{imagenet}  \\ 
Robo-VQA-E  & Re-organize & Real-world & 350 & 5.5 & 4.1 & ROM~\cite{robovqa}   \\ 
\specialrule{1.5pt}{0pt}{0pt}
\end{tabular}
}
\label{tab:datastats}
\end{table*}
\begin{figure*}[!t]
\centering
\begin{subfigure}{.485\textwidth}
  \centering
\includegraphics[width=0.95\linewidth]{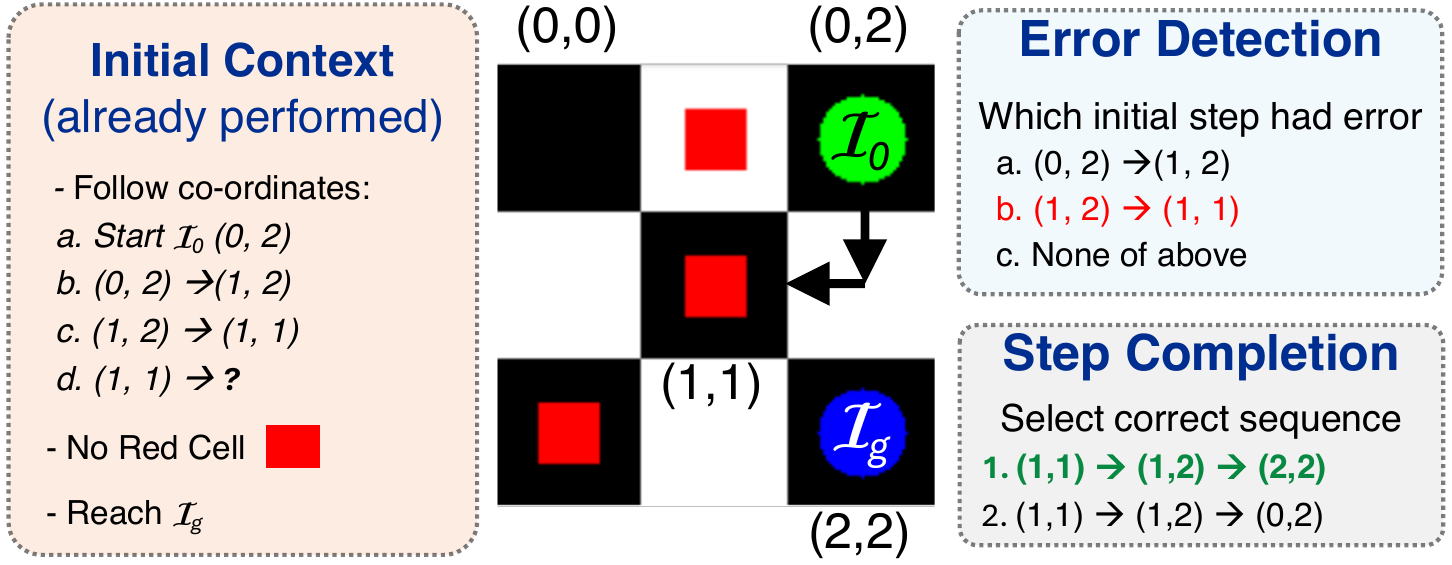}
\caption{\textbf{Maze-E}: Only navigating ($\rightarrow, \uparrow, \leftarrow, \downarrow$) from green  to blue cell, avoiding cells marked in red. }
\vspace{5pt}
\label{fig:Maze-E}
\end{subfigure}%
\hfill
\begin{subfigure}{.485\textwidth}
  \centering
\includegraphics[width=0.95\linewidth]{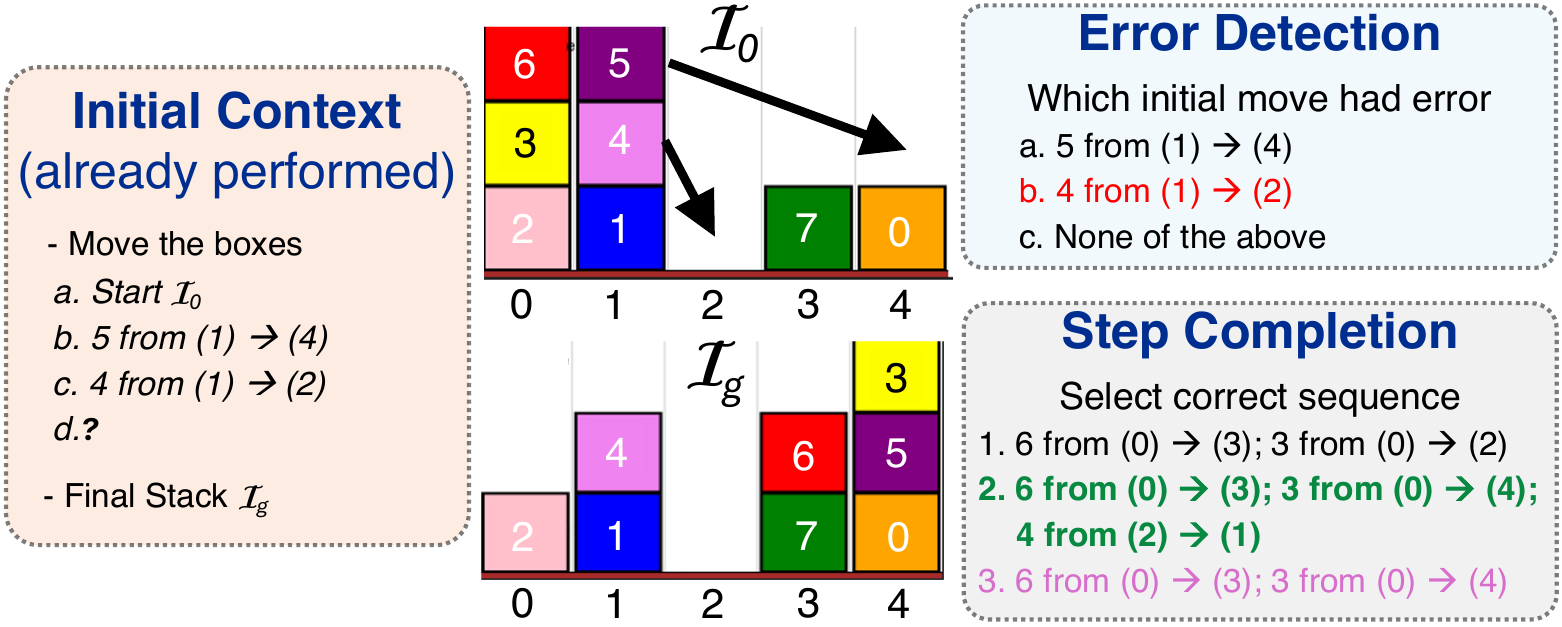}
\caption{\textbf{Blocks-World-E}:
X from $(a)\rightarrow(b)$ re-arranges block `X' from column `a' to column `b'. 
}
\vspace{5pt}
\label{fig:Blocks-World}
\end{subfigure}
\hfill
\begin{subfigure}{.49\textwidth}
  \centering
\includegraphics[width=\linewidth]{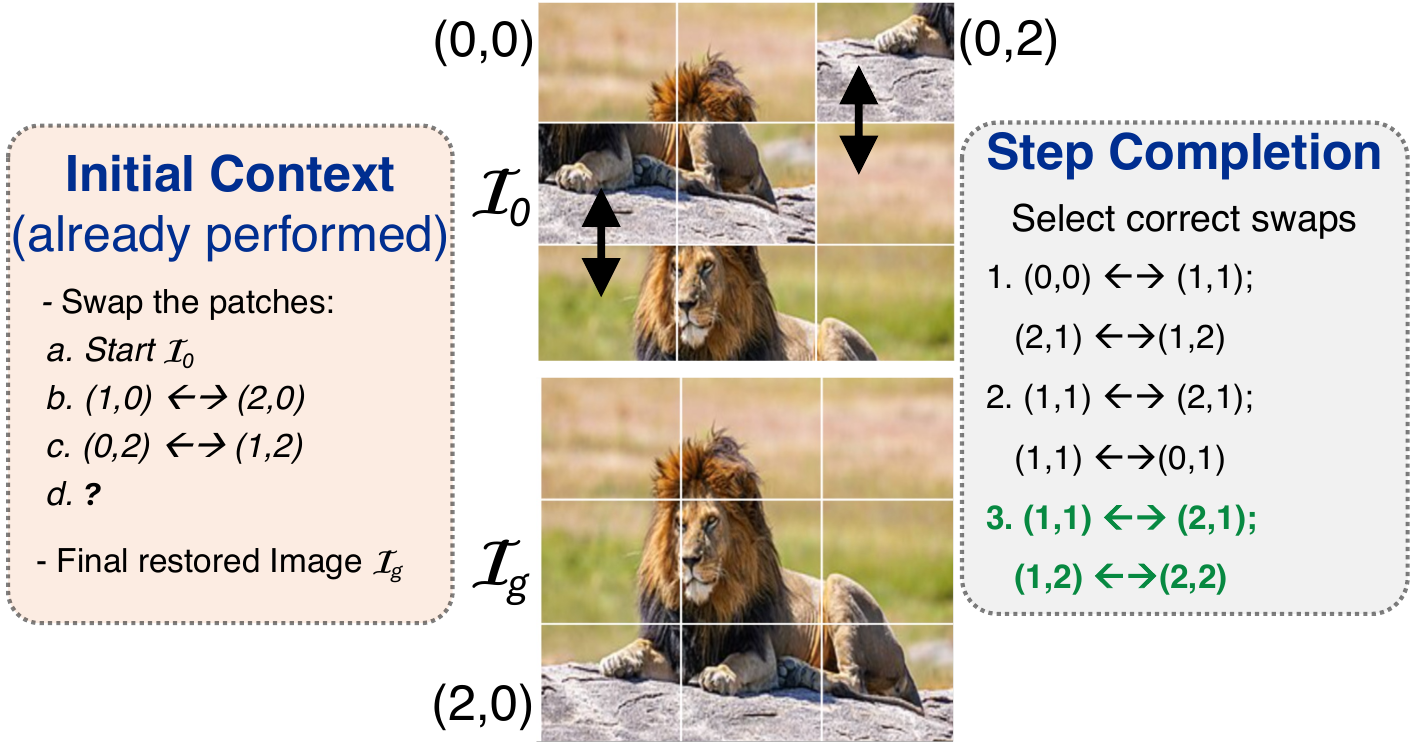}
\caption{\textbf{Shuffle-E}: Re-constructing target image $\mathcal{I}_g$ by swapping ($\leftrightarrow$) image patches.}
\label{fig:Shuffle-E}
\end{subfigure}
\hfill
\begin{subfigure}{.49\textwidth}
  \centering
\includegraphics[width=\linewidth]{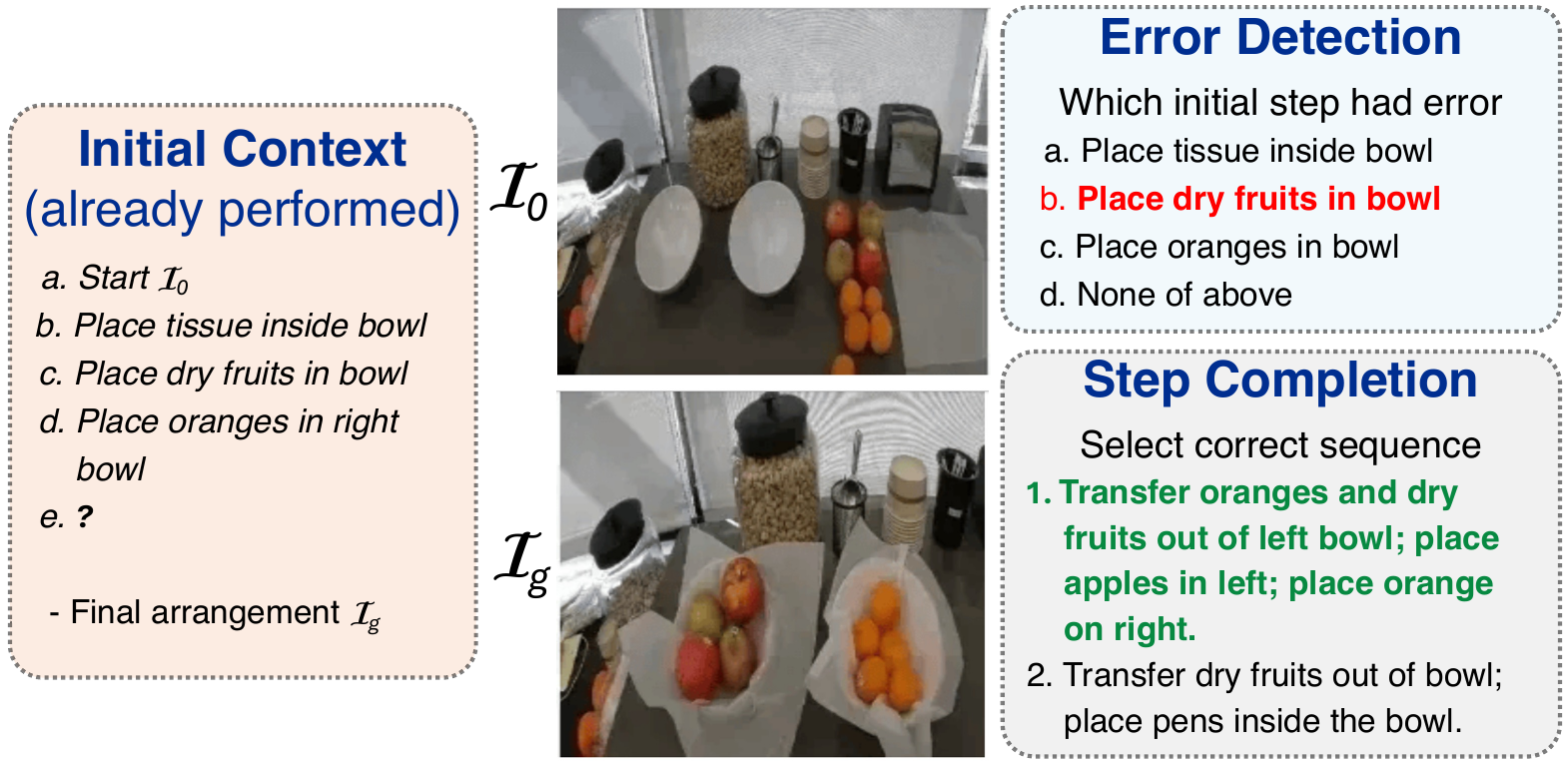}
\caption{\textbf{Robo-VQA}: Re-organizing real-world objects.}
\label{fig:Robo-VQA}
\end{subfigure}

\caption{\textbf{\CPP}: 
We provide the initial ($\mathcal{I}_0$) and the final state ($\mathcal{I}_g$), with an initial set of already performed actions (initial context in orange, visualized as black arrows).
We introduce \textit{one error} that either violates the rules of the environment or is sub-optimal; however, one error in Shuffle-E would have a cascading effect, requiring multiple backtracks for correction, hence ignored (\cref{sec:tasks}).
VLMs perform: \textit{Step Completion}, choose the correct set of future steps (green) to reach $\mathcal{I}_g$, and \textit{Error Detection}, detecting the erroneous step (red) in the initial context. 
Coordinates (row, column) are 0-indexed. 
Safeguard against cheating (\cref{sec:safeguard}) highlighted in pink for Blocks-World-E.
}
\label{fig:taskoverview}
\end{figure*}

\section{\CPP Benchmark}
\label{sec:benchmark}

\underline{\textbf{Co}}rrective \underline{\textbf{S}}equence \underline{\textbf{Plan}}ning 
(\textbf{\CP}) 
mimics general decision-making by evaluating the model’s ability to navigate a complex challenge of detecting and correcting non-optimal (error) steps in a sequential planning task. 
In this setup, 
$\bullet$ Model $\mathcal{M}$ progresses from an initial state $\mathcal{I}_0$ to a goal state $\mathcal{I}_g$ through a sequence of \textit{N} actions:
($\mathcal{A}_1,\mathcal{A}_2,...\mathcal{A}_N$). 
$\bullet$ We introduce an intentional non-optimal (error) action  
\Er within the \textit{initial context (already performed $k$ actions)} 
($\mathcal{A}_1,\mathcal{A}_2,..\MER..\mathcal{A}_{k<N}$).
$\bullet$ The model must \textit{detect} this erroneous action \Er and course-correct to \textit{complete} the remaining actions steps ($\mathcal{A}_{k+1},\mathcal{A}_{k+2},...\mathcal{A}_{N}$) towards the final goal.
Mathematically, it can be shown as 

\begin{align}
\begin{array}{l}
    \text{\textbf{\CP}}\\
   \mathcal{M}( \mathcal{A}_{1,..\EER,..k}; \mathcal{I}_0; \mathcal{I}_g) \\ 
   \text{Initial State: } \mathcal{I}_0 \\ 
   \text{Goal: } \mathcal{I}_g  \\ 
   \text{Performed actions: } 
   \mathcal{A}_{1,..\EER,..k}
   \end{array} & \rightarrow 
\begin{cases}
   \text{\textbf{Error Detection}} \\
   \text{Identify} \MER  \vspace{6pt} \\
   \text{\textbf{Step Complete}} \\
   \mathcal{A}_{k+1,k+2,..N} 
\end{cases}
\end{align}

This setup is used to solve diverse scenarios, such as \textit{re-constructing} a correct image from shuffled image tiles, \textit{re-arranging \& re-organizing} the objects / blocks into a coherent order (obeying physics), and 
\textit{navigating} through a maze.
Success relies on addressing and resolving non-optimal (errors) steps encountered along the way (\figno{\cref{fig:taskoverview}}). 
Sec. \ref{sec:tasks} describes each dataset with proposed planning tasks.

\vspace{4pt}
\noindent \textbf{What's an Error?}
We loosely define `error' as a plausible but suboptimal action that deviates from the optimal path to the goal, potentially resulting in longer sequences. An error can also be a purely wrong action that makes it impossible to reach the goal without correction, \eg referencing non-existent objects, violating task/physics constraints \etc 
Presence of erroneous action forces models to reason over \textit{all actions / intermediate states} to identify and correct the error.

\subsection{Benchmark Datasets}
\label{sec:tasks}

We introduce four sequential planning datasets, each featuring diverse tasks with intentional sub-optimal errors (except `Shuffle-E') posing unique challenges in corrective sequence planning.
Similar to previous planning benchmarks~\cite{spatial2024eval,grover2024navigatinghallucinationsreasoningunintentional,Tran_2025_ICCV,fan2025vrexbenchmarkingexploratoryvisual}, we opt for multiple-choice questions~\cite{NEURIPS2024_89cc5e61} to test:
i) \textbf{Error Detection}: Identifying the non-optimal erroneous action from initial context (already performed actions), or selecting ``none of the above".
ii) \textbf{Step Completion}: Selecting the correct answer among 5 options that would correct the mistake and lead to the final goal.
The use of synthetic datasets~\cite{wang2024pictureworththousandwords,pothiraj2025capture} has been shown to test reasoning vulnerabilities in VLMs.
An overview is provided in \cref{tab:datastats} \& \cref{fig:taskoverview}, respectively.
Intuition behind this formulation explained in \SUPP.

\vspace{4pt}
\noindent \textbf{Maze-E} 
(\figno{\cref{fig:Maze-E}}):
The \textit{goal} is to solve a maze while navigating from the start cell (green, $\mathcal{I}_0$) to the goal cell (blue $\mathcal{I}_g$). 
\textit{Input} is a maze layout with $\mathcal{I}_0$, $\mathcal{I}_g$, initial sequence of moves under the constraints of moving \textit{only} up, down, left, or right, within maze boundaries, while \textit{avoiding} red cells. 
Dataset is constructed via randomly sampled grid of size $\in [3\times 3, 8\times 8]$, and up to 5 obstacles\footnote{Black \& white pattern helps distinguish cells/navigate, using OpenCV}.

\vspace{4pt}
\noindent \textbf{Blocks-World-E} (\figno{\cref{fig:Blocks-World}}):
It's a block rearrangement task where the objective is to transform an initial configuration ($\mathcal{I}_0$) into a target configuration ($\mathcal{I}_g$).
\textit{Input} includes $\mathcal{I}_0$, $\mathcal{I}_g$ arrangements along with a partially executed stacking sequence under the constraint of moving \textit{only} top blocks (nothing on top) and place it either on another top block or empty column. 
The dataset is generated using OpenCV, with 3 to 8 blocks randomly distributed across columns.

\vspace{4pt}
\noindent\textbf{Shuffle-E} (\figno{\cref{fig:Shuffle-E}}):
Here, the objective is to recover (reconstruct) the original image by swapping shuffled image patches.
\textit{Input} is a starting shuffled image ($\mathcal{I}_0$), the final restored image ($\mathcal{I}_g$), and the initial sequence of image patch swaps.
Dataset contains 1000 uniformly sampled ImageNet~\cite{imagenet} images across all classes.



\vspace{4pt}
\noindent \textbf{Robo-VQA-E} (\figno{\cref{fig:Robo-VQA}}):
Goal is to perform a sequence of actions to re-organize real-world objects in an initial scene ($\mathcal{I}_0$) to reach desired arrangement ($\mathcal{I}_g$).
The dataset (modified ROM~\cite{robovqa}) contains 350 curated image-text pairs. 
\textit{Inputs}: starting scene, target scene, and a sequence of initial object placement actions.

\begin{figure}[!t]
\centering
\centering
\includegraphics[width=\linewidth]{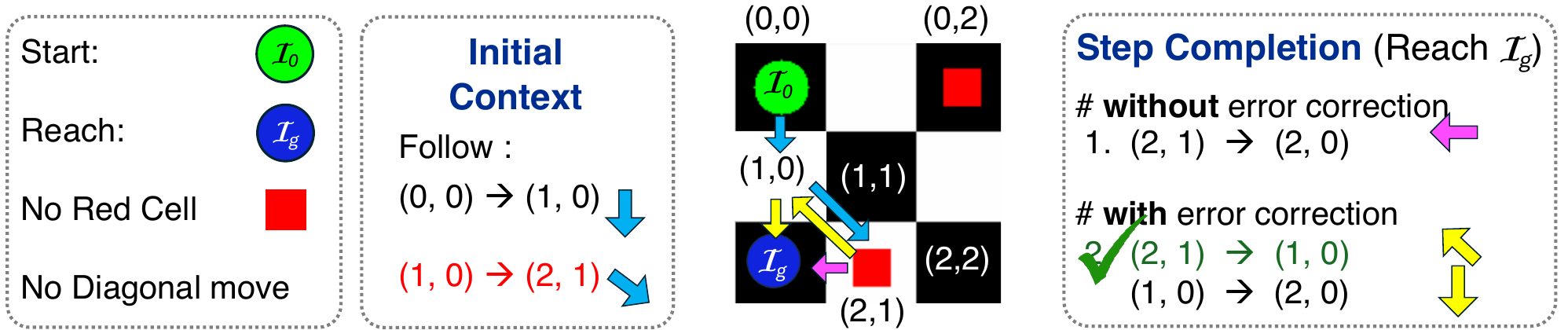}
\caption{
\textbf{Error Correction:} \BLUE{Initial context} has \RED{erroneous} move from $(1,0)$ to red  cell $(2,1)$. 
True option corrects this error (yellow), while the one without (pink) isn't correct. 
}
\label{fig:errorcorrstep}
\end{figure}

\subsection{Erroneous Step \Er} 
We introduce one error step in the initial context.
For \textbf{Maze-E}, we randomly select one of the following incorrect moves: i) move out of bounds, ii) move into an obstacle (red cell), and iii) move diagonally.  
For \textbf{Blocks-World-E} error involves randomly picking \textit{one block} and randomly picking one of the wrong columns, either i) stacking inefficiently ii) placing blocks in physically impossible positions, such as floating in the air or between blocks, or iii) moving non-top blocks.
In \textbf{Robo-VQA-E}, we use Qwen 2 VL-8B (curated by us) to annotate the scene and pick two random objects/containers, \eg pen, dustbin, to generate errors.
These include: 
1) Suboptimal rearrangement \eg ``\textit{place obj1 }(pen) \textit{in cont2} (dustbin)"
2) Arbitrary interaction with objects outside the scene \eg ``\textit{break obj1 using obj2}"
3) Arbitrary or incorrect placements, ``\textit{cover cont1 "}.
3) Invalid actions \eg ``\textit{opening an already open door''}. For \textbf{Shuffle-E} an erroneous patch swap will cascade into subsequent wrong swaps, making it \textit{infeasible} to maintain uniformity across other one-error settings.

\subsection{Safeguard against Shortcuts and Cheating 
}
\label{sec:safeguard}
Revealing the final scene $\mathcal{I}_g$ enables shortcutting: the model can simply choose the option that matches $\mathcal{I}_g$ without reasoning about the initial state or constraints. To prevent this, we add a distractor that also reaches $\mathcal{I}_g$ but \textbf{omits} the required error-correction step.
 Conversely, correct selection requires understanding the scene to rectify actions, rather than relying on final-state visual similarity. \figno{\cref{fig:Blocks-World}} `Step Completion' shows option 2 (green) and option 3 (pink) both match $\mathcal{I}_g$, but only option 2 corrects the error of moving block 4 from the 2\textsuperscript{nd} column to 1\textsuperscript{st} column. 
Similarly, in \figno{\cref{fig:errorcorrstep}}, the option 2 (yellow) reverts the illegal diagonal move, 
while option 1 simply reaches the target.

\begin{figure}[!t]
\centering
\includegraphics[width=0.495\linewidth]  {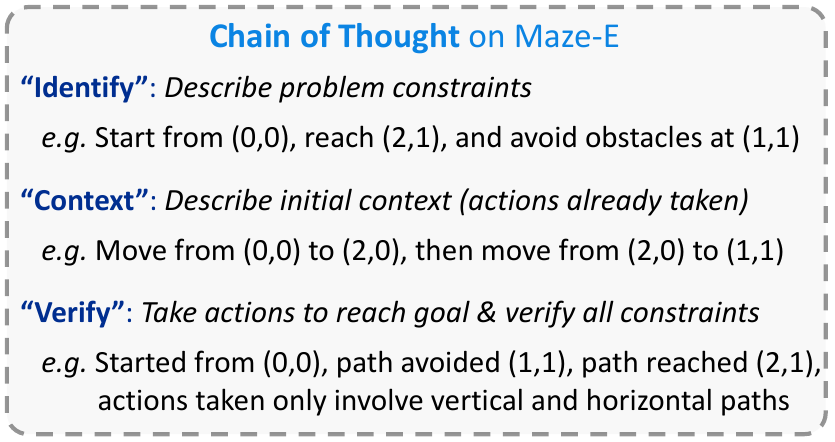}
\hfill
\includegraphics[width=0.495\linewidth]  {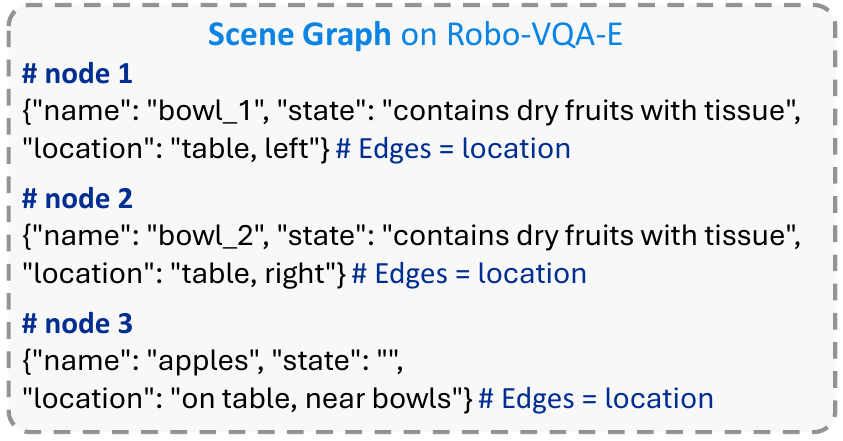}
\caption{\textit{(Left)\textbf{Chain of Thought}}: All VLMs are asked to iterate actions step-by-step (context) and verify if they reach the goal.
\textit{(Right) \textbf{Scene Graph}}: 
 GPT-4o describes objects as \textit{nodes}, location as \textit{edges}, state as \textit{attributes}. 
Detailed description in~\cref{sec:cot_reason}.
}
\label{fig:cot_ex}
\label{fig:scene_graph_ex}
\end{figure}

\subsection{Evaluation}
\label{sec:eval}
\CPP is evaluated using a multiple-choice question (MCQ) framework (shown in \cref{fig:taskoverview}), with Top-1(\%) accuracy as the evaluation metric\footnote{Higher number implies better performance $\uparrow$.}.
We \textit{independently} evaluate \textbf{Step Completion:} the model chooses \textit{one correct option from 5} options (random accuracy $\frac{1}{5}$), and 
\textbf{Error Detection:} MCQ setup presents \textit{initial context actions as choices}, with an additional option of `none of the above' denoting no error present (random accuracy $= \ER [ \frac{1}{\text{Initial context length} + 1}] $). 

\subsection{Models \& Techniques}
\label{sec:cot_reason}
\label{sec:scene_graph}

Following the OpenVLM leaderboard and the recent \textit{Capture}~\cite{pothiraj2025capture} benchmark, we evaluate \CPP on GPT-5.1, GPT-4o, CoG-VLM, Intern-VLM 2, Intern-VLM 3, Janus-Pro-7B, Llama3-8B, Qwen2-VL-8B, and Qwen3-VL-8B. As GPT-4o and GPT-5.1 are paid models, we use them for limited experiments.
We further examine enhanced reasoning strategies Chain-of-Thought (CoT)~\cite{wei2022chain} and Scene Graphs (SG)~\cite{cot_reasoning_vlm2023} to improve VLM performance. Detailed comparisons between SG and CoT, along with model specifics, are provided in \SUPP.

\vspace{6pt}
\noindent \textbf{Chain-of-Thought (CoT)}
CoT~\cite{wei2022chain, shao2024visualcotadvancingmultimodal} has 3 steps i) \textit{Identify:} 
VLMs are provided with \textit{ `additional'} details of the problem and constraints;
ii) \textit{Context:} Step-by-step description (enforcing) of each action (from the initial context);
iii) \textit{Verify:} \textit{`additional'} verification that the VLMs follow all constraints and steps. 
This approach is model-agnostic, and the same set of instructions is provided to all models. An example is shown in \cref{fig:cot_ex} (left) and in \SUPP. 

\vspace{6pt}
\noindent \textbf{Scene Graphs (SG)}
SG adds spatial awareness to CoT, aiding reasoning abilities. We \underline{`QUERY'} (prompt) VLM to
construct SG for the initial ($\mathcal{I}_0$) and goal states ($\mathcal{I}_g$), capturing three key components: a) Objects in the scene (\textit{nodes}), b) Attributes (\textit{properties}) and interactions between objects, and c) Spatial relationships (\textit{edges}) capturing relative positions.
An example SG is shown in \cref{fig:scene_graph_ex} (right) and \SUPP.
Different VLM's have different structural representations (model-specific SG), however, we ensure uniform prompt for all VLMs.

\begin{table*}[!t]
\centering
\caption{\textbf{\CPP Step Completion:} 
`V' denotes vanilla (no CoT or SG). Both initial \& final SG(s) are input to VLM.
$\dag$ means vanilla evaluation skipped because of financial constraints (CoT \& SG generally outperform vanilla anyways); 
\RCTEXT{Lighter shade} indicates $\pm 2\%$ close to random choice; \WRTEXT{Darker} is intentionally choosing wrong ($<$ random).
}
\renewcommand{\arraystretch}{0.95}
\setlength\tabcolsep{2pt}
\resizebox{\textwidth}{!}{
\begin{tabular}{l | ccc | ccc | ccc |ccc}
\specialrule{1.5pt}{0pt}{0pt}
\rowcolor{mygray}
& 
\multicolumn{3}{c|}{\textbf{Robo-VQA-E (\%)}} & 
\multicolumn{3}{c|}{\textbf{Shuffle-E (\%)}} & 
\multicolumn{3}{c|}{\textbf{Maze-E (\%)}}
 &
\multicolumn{3}{c}{\textbf{Blocks-World-E (\%)}} \\
[-.4pt]\hhline{~ --- --- --- ---}
\rowcolor{mygray}
\multirow{-2}{*}{ \textbf{VLM} } & 
V & CoT & SG & 
V & CoT & SG & 
V & CoT & SG & 
V & CoT & SG \\ 
\hline
Random & \multicolumn{3}{c|}{ \RC{20}} & \multicolumn{3}{c|}{\RC{20}} & \multicolumn{3}{c|}{\RC{20}}  & \multicolumn{3}{c}{\RC{20}} \\ 
Human & \multicolumn{3}{c|}{42.1} & \multicolumn{3}{c|}{53.7} & \multicolumn{3}{c|}{95.7}  & \multicolumn{3}{c}{81.8} \\ 

\hline
Llama3-8B~\cite{grattafiori2024llama3herdmodels} & 
\RC{18.1} & \RC{18.3} & \RC{19.1} & 
\WR{17.3} & \WR{17.7} & \RC{18.5} & 
\RC{19.5} & \RC{20.1} & \RC{21.3} & 
\RC{21.3} & 22.7 & 23.2 \\ 

CoG-VLM~\cite{cogvlm} & 
\WR{13.1} & \WR{12.5} & \RC{21.5} & 
23.1 & \HB{27.1} & 23.7 & 
25.1 & 25.9 & 26.5 & 
25.5 & 25.2 & 26.7 \\ 

Janus-pro-7B~\cite{janus} & 
\WR{14.1} & \WR{14.7} & \RC{21.3} & 
23.2 & 23.1 & 23.5 & 
\RC{20.4} & \RC{20.2} & \RC{21.7} & 
24.2 & 23.1 & 25.1 \\ 
Qwen2 VL-8B~\cite{qwen} & 
\WR{17.1} & \WR{17.6} & \RC{18.9} & 
24.1 & 24.9 & 25.1 & 
26.5 & 27.9 & 28.3 & 
\RC{18.1} & \RC{18.6} & \RC{18.8} \\ 

Qwen3 VL-8B~\cite{bai2025qwen3vltechnicalreport} & 
\RC{20.3} & \RC{21.2} & 23.4 & 
28.3 & 29.4 & 30.3 & 
35.1 & 34.6 & 36.2 & 
26.5 & 27.9 & 28.3 \\ 
Intern-VLM 2~\cite{internvl} & 
22.1 & 23.5 & 25.1 & 
\RC{20.1} & 23.2 &  23.4 & 
\RC{21.6} & \HB{35.8} & \HB{41.2} &
\RC{18.3} & \RC{21.2} & \RC{18.9} \\

Intern-VLM 3~\cite{wang2025internvl35advancingopensourcemultimodal} & 
27.1 & 29.4 & 31.4 & 
24.1 & 25.6 &  27.1 & 
48.6 & \HB{48.8} & \HB{50.1} &

25.7 & 27.1 & 29.4 \\ 

\hline 
GPT-4o~\cite{gpt4} $\dag$& 
- & 48.2 & \HB{52.2} & 
- & 27.6 & 30.1 & 
- & 45.6 & 46.1 & 
- & 49.7 & \HB{54.3} \\
GPT-5.1~\cite{gpt4} $\dag$& 
- & 49.3 & \HB{51.3} & 
- & 31.5 & 33.4 & 
- & 49.1 & 48.3 & 
- & 48.1 & \HB{52.3} \\

 Gemini-3-pro~\cite{team2023gemini} $\dag$& 
 - & 61.6 & \cellcolor{green!20}\HB{67.3} & 
 - & 57.4 &\cellcolor{green!20} 61.6 & 
 - & 68.2 & \cellcolor{green!20}70.4 & 
 - & 69.5 & \cellcolor{green!20}\HB{71.3} \\ 
\specialrule{1.5pt}{0pt}{0pt}
\end{tabular}
}
\label{tab:results}
\label{tab:main_abla}
\end{table*}

\begin{table*}[!t]
\centering
\caption{\textbf{\CPP Error Detection:} 
Format same as \cref{tab:results}. Models look at the initial context and try to identify the erroneous action.}
\renewcommand{\arraystretch}{0.95}
\setlength\tabcolsep{7pt}
\resizebox{\textwidth}{!}{
\begin{tabular}{l | ccc | ccc | ccc}
\specialrule{1.5pt}{0pt}{0pt}
\rowcolor{mygray}
& 
\multicolumn{3}{c|}{\textbf{Robo-VQA-E}} & 
\multicolumn{3}{c|}{\textbf{Maze-E}} &
\multicolumn{3}{c}{\textbf{Blocks-World-E}}  \\ 
[-.4pt]\hhline{~ --- --- ---}
\rowcolor{mygray}
\multirow{-2}{*}{ \textbf{VLM} } & 
V & CoT & SG & 
V & CoT & SG & 
V & CoT & SG  \\ 
\hline
Random & \multicolumn{3}{c|}{\RC{25.4}} & \multicolumn{3}{c|}{\RC{26.1}} & \multicolumn{3}{c}{{26.1}} \\ 
Human & \multicolumn{3}{c|}{60.0} & \multicolumn{3}{c|}{90.0} & \multicolumn{3}{c}{80.0} \\
\hline
Llama3-8B~\cite{grattafiori2024llama3herdmodels} & 
\WR{11.2} &  \WR{10.7} & \WR{13.6} & 
\WR{18.7} & \WR{18.3} & \WR{19.7} & 
\RC{27.3} & \RC{27.6} & \RC{28.2} \\ 

CoG-VLM~\cite{cogvlm} & 
32.1 & 33.4 & 35.3 & 
\WR{6.4} & \WR{8.4} & \WR{13.3} & 
41.3 & 43.1 & 44.5 \\ 

Janus-pro-7B~\cite{janus} & 
\WR{17.5} & \WR{18.1} & \RC{26.1} & 
\WR{20.5} & \WR{19.1} & \WR{21.0} &
29.3 & 31.0 & \RC{27.6} \\

Qwen2 VL-8B~\cite{qwen} & 
\WR{9.2} &  \WR{9.1} & \WR{9.6} & 
\WR{20.5} & \WR{20.8} & \WR{20.7} & 
32.3 & 30.6 & 35.2 \\ 

Qwen3 VL-8B~\cite{bai2025qwen3vltechnicalreport} & 
\WR{13.1} &  \WR{15.6} & \WR{18.1} & 
\RC{23.8} & \RC{24.1} & \RC{25.8} & 
34.1 & 35.3 & 36.7 \\ 

Intern-VLM 2~\cite{internvl} & 
\RC{24.3} & \RC{25.2} & \RC{26.1} & 
32.8 & 33.1 & 33.4 & 
36.5 & 37.9 & 37.3 \\ 

Intern-VLM 3~\cite{wang2025internvl35advancingopensourcemultimodal} & 
\RC{25.1} & \RC{26.6} & \RC{28.1} & 
33.3 & 34.7 & 35.1 & 
42.4 & 43.1 & 44.3 \\ 

\hline
GPT-4o~\cite{gpt4} $\dag$& 
- & 45.3 & 44.2  & 
- & 40.3 & \HB{35.3} & 
- & 35.1 & \HB{42.1} \\

GPT-5.1 \cite{gpt4} $\dag$& - & 54.9 & 46.3 & - &39.8 & 37.7 & - & 38.3 & 44.6 \\
 Gemini-3-pro~\cite{team2023gemini} $\dag$& 
 - & 57.3 & \cellcolor{green!20}62.5  & 
 - & 62.6 & \cellcolor{green!20}\HB{67.8} & 
 - & 67.5 &\cellcolor{green!20} \HB{71.8} \\

\specialrule{1.5pt}{0pt}{0pt}
\end{tabular}
}

\label{tab:main_abla_error}
\end{table*}

\subsection{Results \& Benchmark Analysis}
We conduct our analysis on the \textbf{Step Completion} task using CoT unless mentioned otherwise. Key insights are \hl{highlighted}.
Additional analyses and extended discussions are provided in the \SUPP.

\vspace{6pt}
\noindent \textbf{Benchmark:}
\figno{\cref{tab:results} \& \ref{tab:main_abla_error}} compares VLMs on \CP, via Vanilla method (raw image-text input), CoT and SG.
SG outperforms CoT, which in turn outperforms vanilla models (few exceptions), highlighting need for structured representations in visual reasoning. 
Proprietary very large reasoning models like GPT-4o and GPT-5.1 make reasonably informed decisions while \hl{relatively light-weight \& open-source ones \textbf{perform near or below random chance}}, indicating the inability to visualize (imagine, predict) intermediate steps.
Accuracy less than random (dark red) can partially be explained by overwhelmingly picking certain options \cite{zheng2023large} \hl{(Janus and Qwen 2 select `option A' 90\%+ and 75\%+ times)} or cheats to reach goal without error correction; For SG,  \hl{InternVLM-2 cheats 43\%, InternVLM-3 41\%, CoG-VLM 27\%, and Qwen3 35\%}.
Based on human scores, the difficulty for tasks follows:  Robo-VQA-E $>$ Shuffle-E $>$ Blocks-World-E $>$ Maze-E.
The near-random performance is \textit{only} on one error. 
Generating problems with \hl{multiple cascading errors will have exponential complexity, limited by automation (\textit{e.g.} GPT-4o) not equipped to handle even one error}.

\begin{figure*}[!t]
\centering
\begin{subfigure}{.5\textwidth}
\centering
\includegraphics[width=\linewidth]{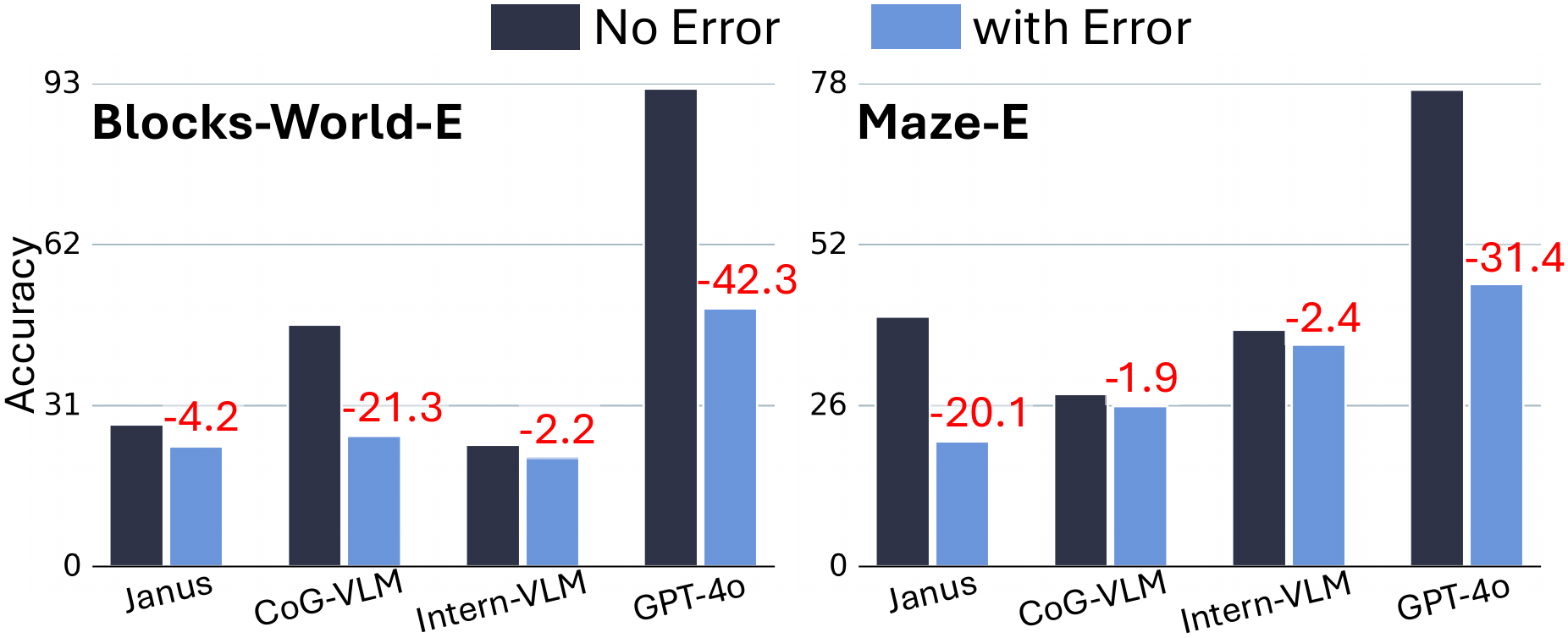}
\end{subfigure}%
\hfill
\begin{subfigure}{.20\textwidth}
\centering
\includegraphics[width=\linewidth]{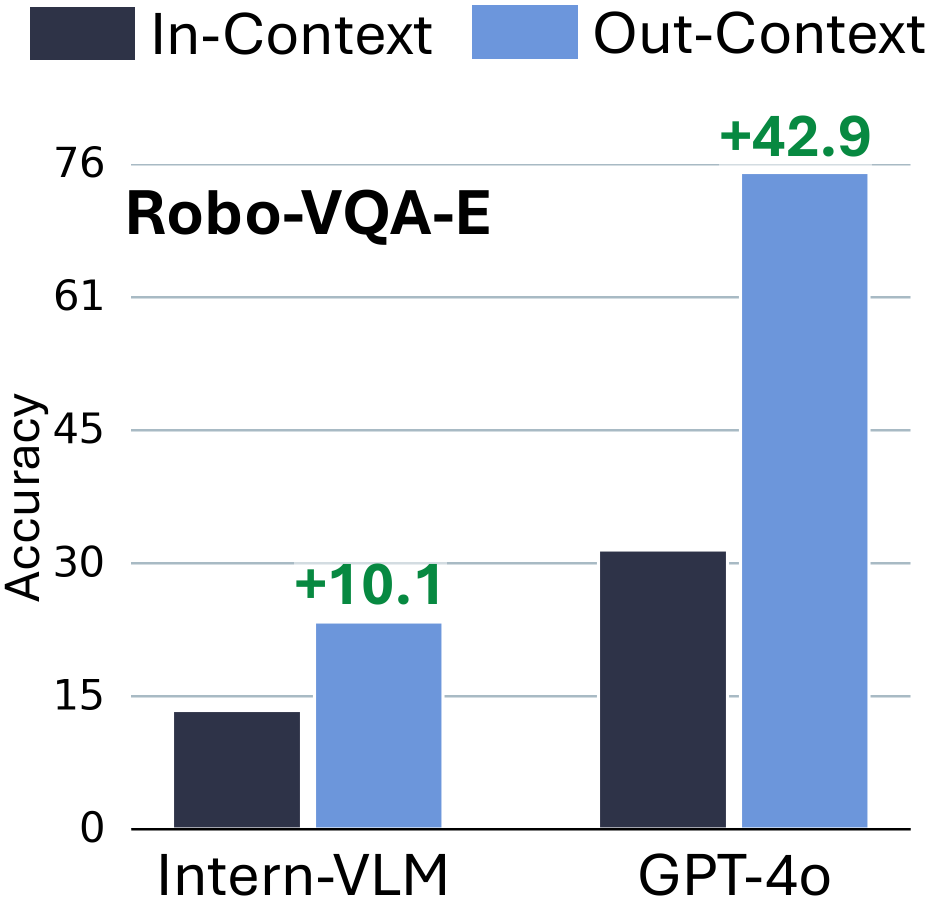}
\end{subfigure}
\hfill
\begin{subfigure}{.27\textwidth}
\centering
\includegraphics[width=\linewidth]{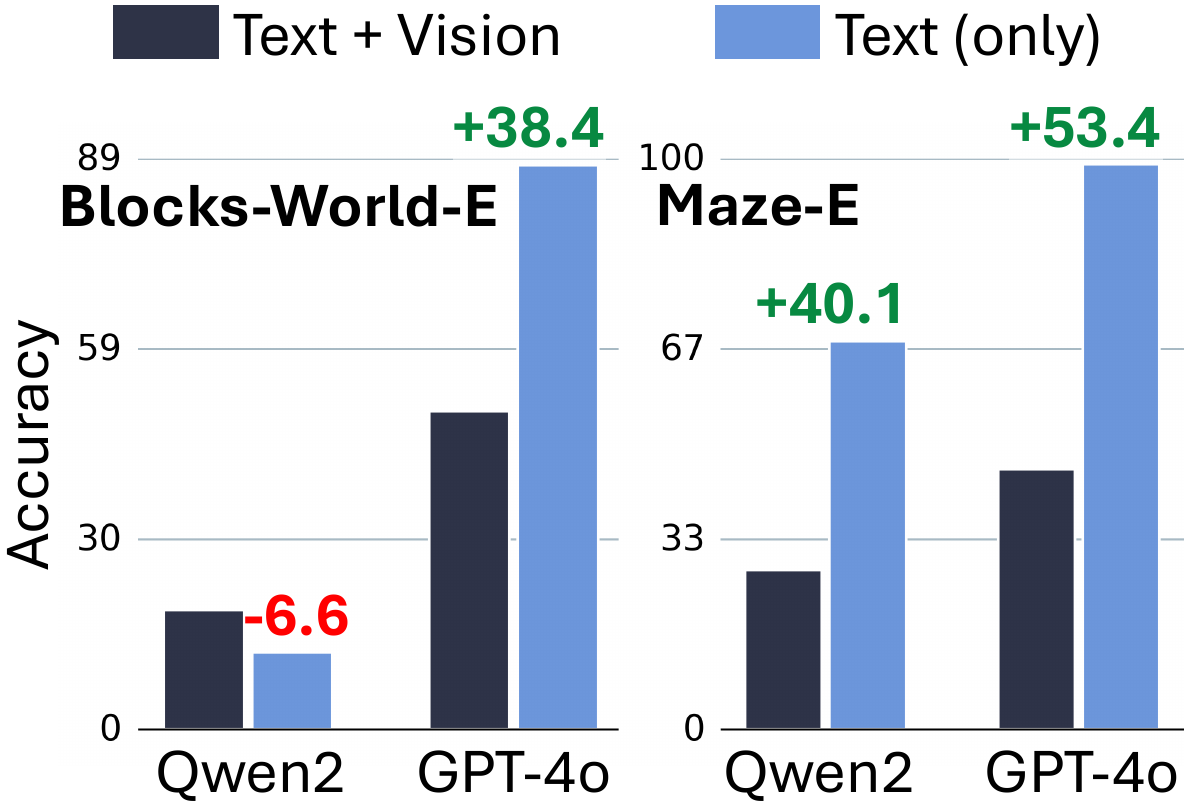}
\end{subfigure}
\caption{
\textit{(Left)} \textbf{Error-free Eval} 
VLMs excel in error-free settings, but struggle on error-prone ones. 
\textit{(Mid)} \textbf{Out vs In-context}: Errors from within context (scene) are harder than random ones (out-context). 
\textit{(Right)} \textbf{Multi-modal vs Text-only } VLMs perform exceptionally well when simulating text-based steps but struggle on visual reasoning. 
}
\label{fig:w_wo_error}
\label{fig:in_context}
\label{fig:fig:w_wo_error}
\label{fig:text_vision}

\end{figure*}

\vspace{6pt}
\noindent \textbf{Impact of Error:}
\figno{\cref{fig:w_wo_error} left)} shows
VLMs excel in error-free settings (GPT-4o near-perfect) but struggle on error-prone ones, hinting at \hl{lack of reasoning, possibly picking options that describes the target} (cheating~\cref{sec:safeguard}).
These highlight the need for challenging, error-prone benchmarks like \CPP to expose the gap between clean training and practical error-prone scenarios, while demanding logical visual reasoning
(CoG-VLM and Janus-pro-7B near random predictions).

\vspace{6pt}
\noindent \textbf{Effect of Context:}
\CPP includes a random mix of two types of errors:
i) \textit{In-Context}: Erroneous step involving objects present in the scene, ii) \textit{Out-Context}: Error uses random objects not in scene (\eg basketball, hammer, \textit{etc}). 
\figno{\Cref{fig:in_context} mid)} shows lower performance on In-Context errors, suggesting \hl{VLMs struggle more when erroneous actions involve \textit{plausible} objects} from within the scene while they can handle the out-of-context errors with relative ease.

\vspace{6pt}
\noindent \textbf{Visual vs Text:}
Similar to previous works~\cite{schulze2025visual,ijcai2025p1164}, describing our tasks (vision+text) in text-only format significantly boosts reasoning accuracy  (\figno{\cref{fig:text_vision} right)}).
This further exposes \hl{models inability to iteratively visualize intermediate reasoning steps, despite performing well in purely text based ones}
(\cref{fig:simplefail}). 
Qwen2's near-random prediction in Blocks-World-E reveals its limitations in logical reasoning. Tasks like Robo-VQA-E and Shuffle-E cannot be faithfully represented as text-only without visual aid. 

\begin{figure*}[!t]
\centering
\begin{subfigure}{.31\linewidth}
\centering
\includegraphics[height=2.45cm]{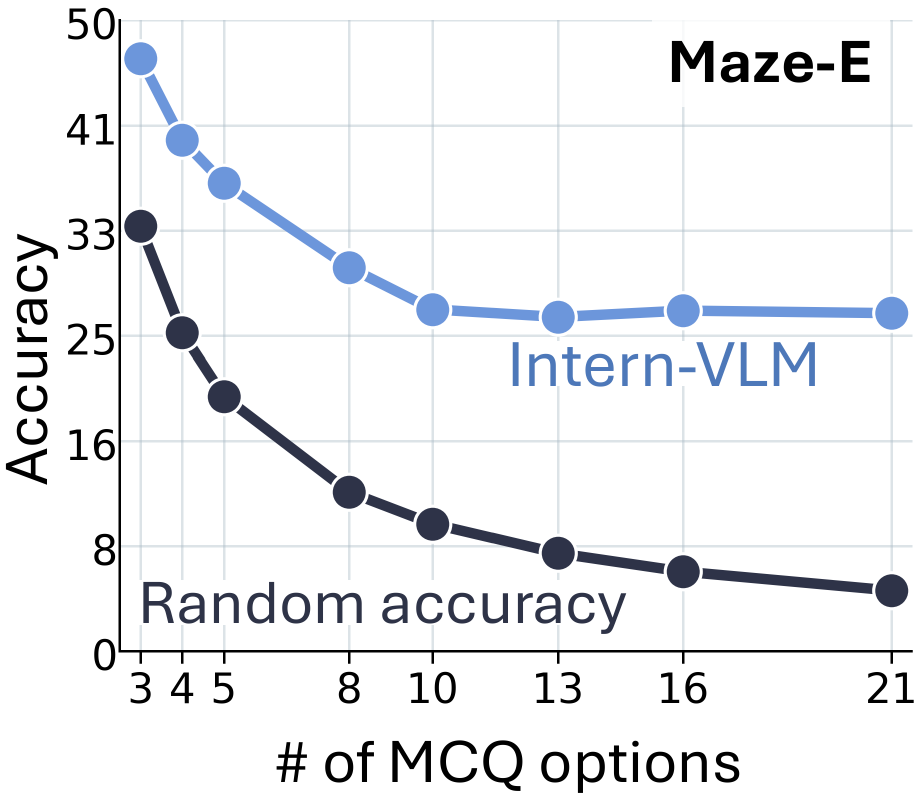}
\end{subfigure}
\hfill
\begin{subfigure}{.34\linewidth}
\centering
\includegraphics[height=2.45cm]{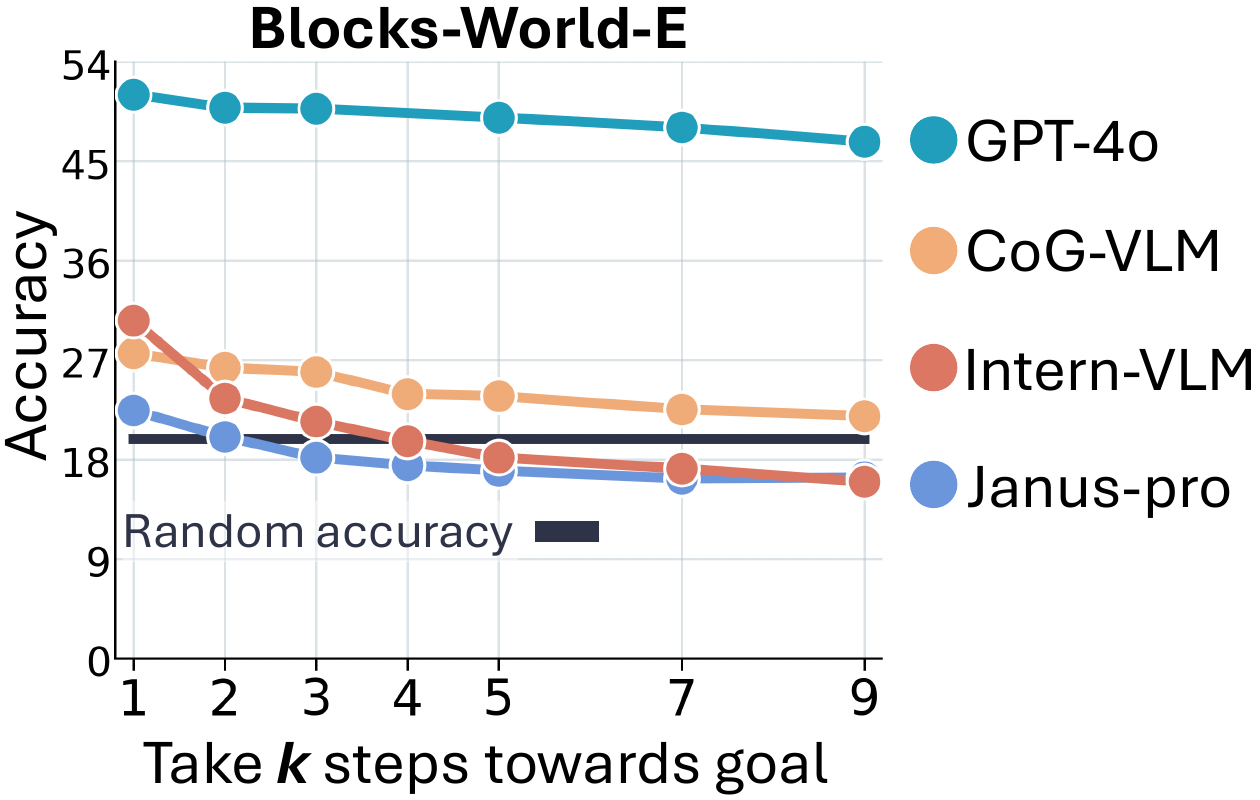}
\end{subfigure}%
\hfill
\begin{subfigure}{.34\linewidth}
\centering
\includegraphics[height=2.45cm]{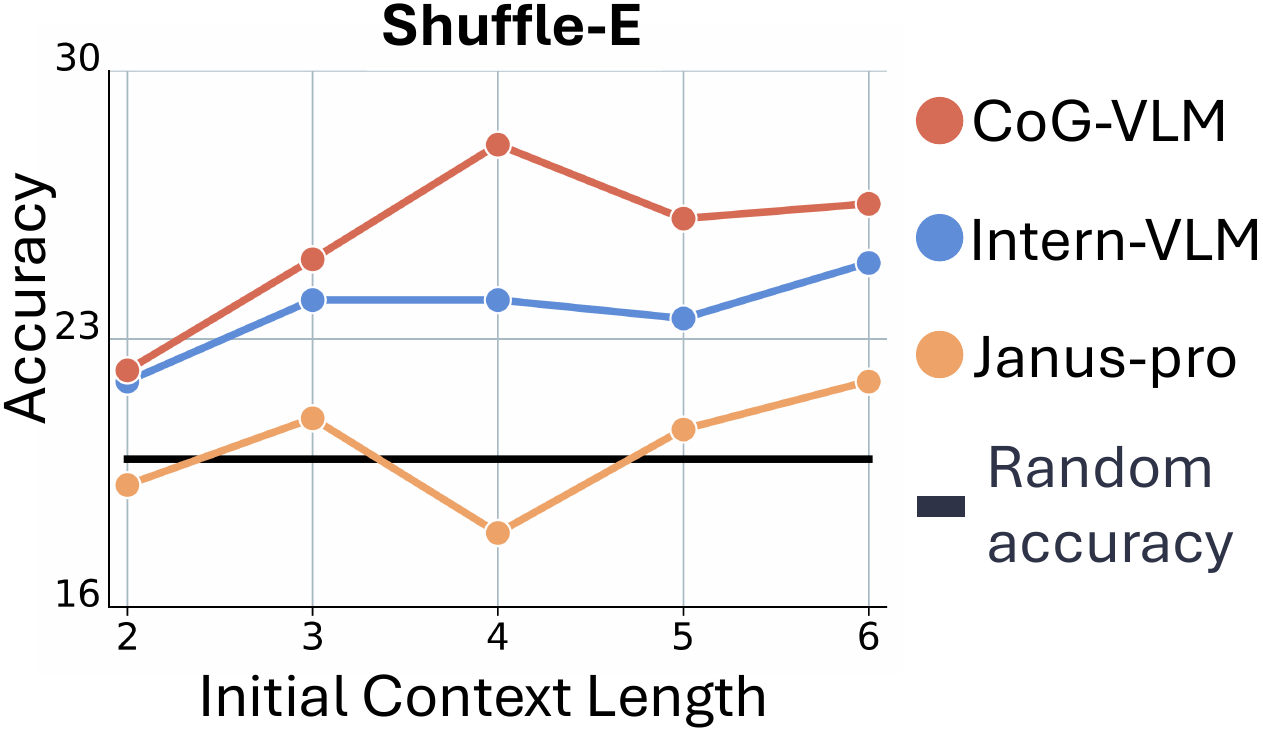}
\end{subfigure}%

\caption{
\textit{(Left)} \textbf{MCQ options count} Intern-VLM (SG) accuracy drops with the \# of MCQ options, hinting at randomness.  
\textit{(Mid)} \textbf{MCQ context}
Step Completion with $9$ remaining steps shows only $k$ steps toward the goal.
Constant accuracy indicates VLMs are not simulating the MCQ option to reach the target. 
\textit{(Right)} \textbf{Length of Initial Context} 
VLM accuracy shows a positive correlation with the number of already performed steps.
}
\label{fig:mcq_options}
\label{fig:k_steps_remain}
\end{figure*}

\vspace{6pt}
\noindent \textbf{MCQ Options:}
\figno{Fig. \ref{fig:mcq_options} left)}
shows that increasing the number of MCQ options drops the performance exposing \hl{the randomness in picking options} (first reported by~\cite{grover2024navigatinghallucinationsreasoningunintentional}).
This reveals the added logical complexity of MCQ in \CP.

\vspace{6pt}
\noindent \textbf{Context from MCQ:}
\figno{\cref{fig:k_steps_remain} mid)} evaluates step completion with 9 remaining steps to reach $\mathcal{I}_g$, under a fixed context length of 2. The MCQ reveals only the next $k$ steps toward $\mathcal{I}_g$ (e.g., $k=2$ exposes only $\frac{2}{9}$ steps and does not reach the goal).All models (including GPT-4o) maintain nearly constant accuracy across $k$, indicating an inability to iterate over visual steps in MCQ options. Ideally, accuracy should improve as more steps toward the goal are revealed. Instead, performance suggests blind option bias (e.g., Janus selects ‘A’ 94\% of the time) and decisions driven solely by the initial context.

\vspace{6pt}
\noindent \textbf{Initial Context importance:}
\figno{\cref{fig:k_steps_remain} right)} shows that model accuracy increases with longer initial context. We hypothesize that including more already-performed, \textit{non-negotiable} steps brings the intermediate state closer to the goal, thereby reducing the burden of predicting the remaining steps.

\vspace{6pt}
\noindent \textbf{Can models visualize actions:}
Given initial, and final state, VLMs are expected to internalize steps between the two states.
\figno{\cref{fig:simplefail}} visualizes how models imagines future steps (states).
Models may succeed on certain steps but fail on subsequent ones, revealing brittle incoherent long-horizon reasoning, partially explaining the random prediction observed on \CP.

\begin{figure}[!t]
    \centering
    \includegraphics[width=0.95\linewidth]{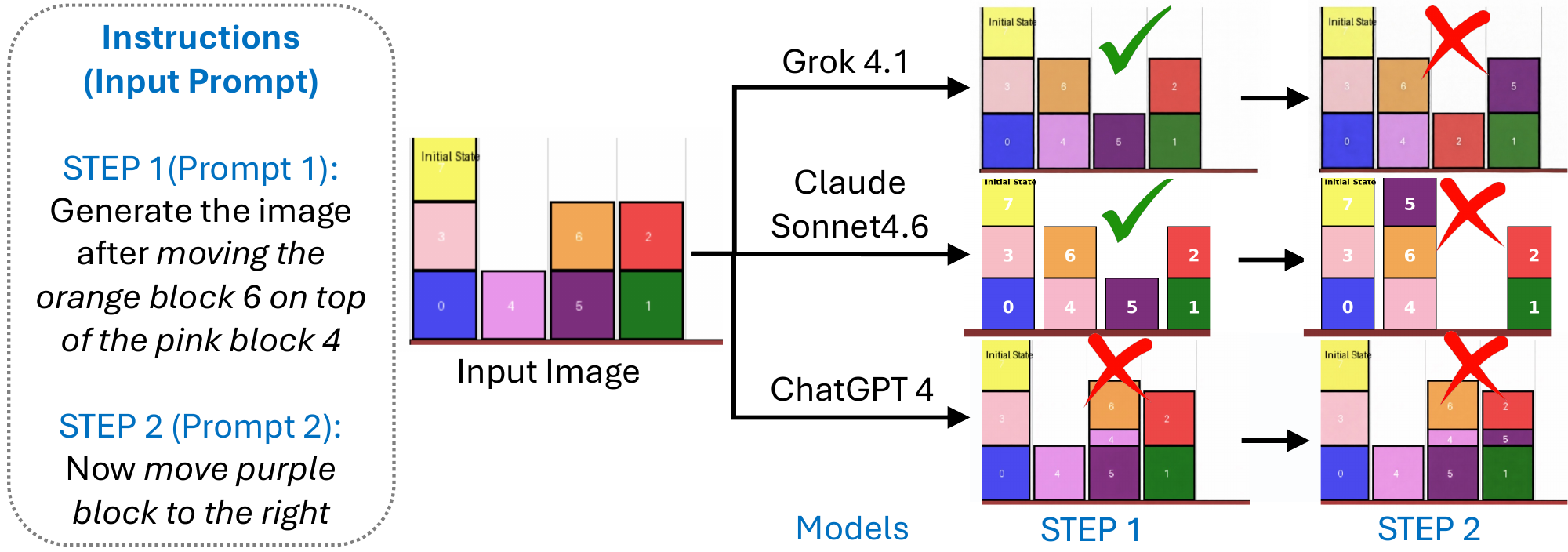}
    \caption{\textbf{Failure in Simulating Visual Steps}:
    Proprietary VLMs like Grok~\cite{xai2025grok4}, Claude~\cite{anthropic2024claude46}, ChatGPT \cite{openai2024gpt4technicalreport} fail to generate intermediate states starting from the initial scene. 
    On Blocks-World-E, Grok misplaces the red block, Claude moves purple block to the left instead of right, and ChatGPT fails to generate correct steps.
    }
\label{fig:simplefail}
\end{figure} 

\begin{figure*}[!t]
\centering
\includegraphics[width=\linewidth]{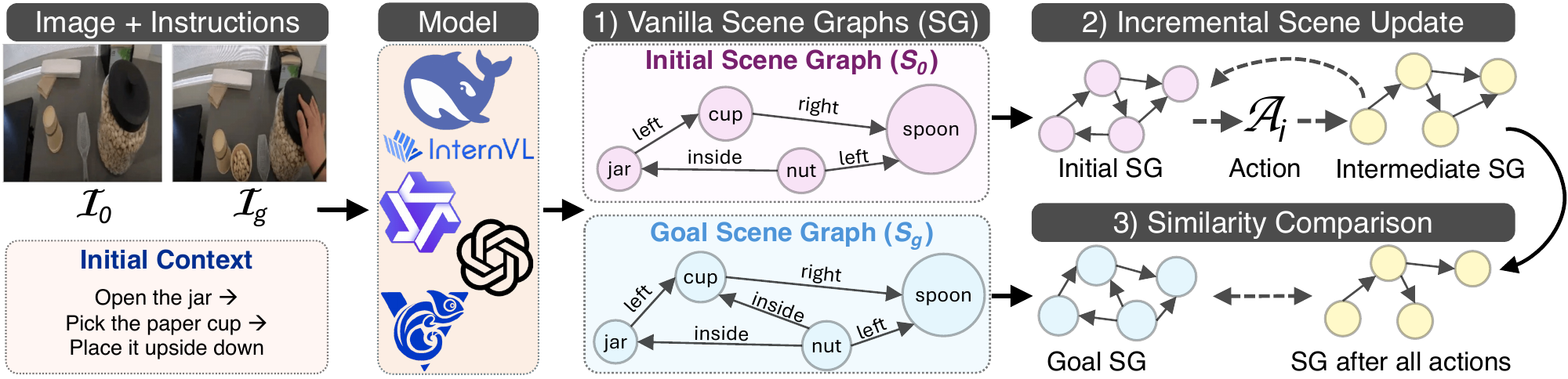}
    \caption{\textbf{SGI:} 1) Initial and Goal Scene Graphs (SG) are generated. 2) Incremental Scene Update sequentially modifies SG for each action $A_i$ 3) Similarity Comparison matches the resultant SG with Goal graph for searching for the best-aligned sequence.
    }
    \label{fig:arch}
\end{figure*}

\section{\underline{S}cene \underline{G}raph \underline{I}ncremental update (SGI)}
\label{sec:sgi}

VLMs near random performance on \CPP (\cref{tab:results}) fails to leverage the intermediate steps (\cref{fig:k_steps_remain} (mid)). 
With only initial and final images, models internally interpolate missing states (CoT \& SG), a process they struggle in the visual domain, even though they perform with relative ease in the text domain (\cref{fig:text_vision} (right)).
Incorporating these, we propose \underline{\textbf{S}}cene \underline{\textbf{G}}raph \underline{\textbf{I}}ncremental update (\textbf{SGI}) to extend Scene Graph reasoning. 
We leverage SG \textit{to transform visual images into textual representations}, and apply an \textit{iterative step-by-step reasoning on text space instead of the visual domain}. 
Instead of reasoning within a single static scene, 
SGI explicitly derives next-time-frame scene graphs as actions unfold in evolving scenes (\figno{\cref{fig:arch}}). 
This incremental formulation bridges the gap between the initial and final states via 
\textit{explicit intermediate states}, and decomposing reasoning into smaller transitions, thereby improving corrective sequence planning and error detection.
The hierarchy can be visualized as 
CoT $\subseteq$ SG $\subseteq$ SGI.

\subsection{Algorithm}
An overview (\figno{\cref{fig:arch}}) and pseudo code for step completion is shown in \figno{\cref{alg:prism_explicit}}. 
More  details and SGI for error detection in \SUPP.

\begin{algorithm}[!t]
{
\caption{\textbf{SGI} for Step Completion (\cref{sec:sgi})}
\label{alg:prism_explicit}
\textbf{Input}: Initial state $\mathcal{I}_0$, Goal state $\mathcal{I}_g$, Initial Context actions $\mathcal{A}_1,\mathcal{A}_2,..\MER..\mathcal{A}_{k<N}$ \\ 
\OBJECT{Best option from MCQ for 
$\mathcal{A}_{k+1},\mathcal{A}_{k+2}...\mathcal{A}_N$ }
\begin{algorithmic}[1]
\REQUIRE VLM $\mathcal{M}$,  Step Completion  $MCQ$ m options. \\ 
 \vspace{5pt}

\COM{1) Initial and goal Vanilla Scene Graph, (ref \Cref{sec:scene_graph})}
\STATE $S_0 \gets \textsc{QUERY}\bigl[ \mathcal{M}(\mathcal{I}_0) \bigr]$  || 
$S_g \gets \textsc{QUERY} \bigl[ \mathcal{M}(\mathcal{I}_g) \bigr]$ 

 \vspace{5pt}
\COM{2) Incremental Scene Update ($S_0 \rightarrow S_c \rightarrow S_m$)}
\STATE $S_c \gets S_0$  
\FOR{ $\mathcal{A}_i$ in [$\mathcal{A}_1,\mathcal{A}_2,..\MER..\mathcal{A}_{k<N}$] } 
    \STATE $S_c \gets \textsc{SIMULATE} \bigl[ \mathcal{M}( S_c, \mathcal{A}_i) \bigr]$ \hfill 
    \COMM{// $\mathcal{M}$ simulates i-th action $\mathcal{A}_i$ to iteratively \\ \hfill update  intermediate context SG $S_c$ } 
\ENDFOR

\vspace{5pt}
\FOR{each option $m \in MCQ$}
 \STATE
$\mathcal{A}^m_{k+1},\mathcal{A}^m_{k+2}...\mathcal{A}^m_{N}$ $\gets m$ || $S_m \gets S_c$ \hfill  \COMM{// actions \& context SG $S_c$ for option m} 

    \FOR{ $\mathcal{A}^m_i$ in [$\mathcal{A}^m_{k+1},\mathcal{A}^m_{k+2}...\mathcal{A}^m_{N}$] }
    \STATE $S_m \gets \textsc{SIMULATE} \bigl[ \mathcal{M}(S_m, \mathcal{A}^m_i) \bigr]$ \hfill 
    \COMM{// iteratively simulate $m^{th}$ option actions}
\ENDFOR
\ENDFOR

 \vspace{3pt}
\COM{3) Similarity : VLM determines similarity between MCQs derived SG $S_m$ and $S_g$}
\STATE $m' \gets \arg\max_{m \in MCQ} \textsc{SIMILARITY} 
\bigl[ \mathcal{M}(S_m, S_g) \bigr]$  
\STATE \textbf{Output:} $m'$
\end{algorithmic}
}
\end{algorithm}

\vspace{4pt}
\noindent \textbf{1) Vanilla Scene Graphs (SG)}: We  \underline{`QUERY'} VLMs to generate the Scene Graphs for the initial state $\mathcal{I}_0$ as $S_0$ and the final goal $\mathcal{I}_g$ as $S_g$ (\cref{sec:scene_graph}). 
We evaluated the performance of these vanilla SGs in~\cref{tab:results} as $SG = [S_0, S_g]$

\vspace{4pt}
\noindent \textbf{2) Incremental Scene Update}: 
Starting from the initial SG ($S_0$), we feed a textual description of each action 
($\mathcal{A}_{1,..\EER,..k}$) to VLM and ask it \underline{`SIMULATE'} the action on the SG producing intermediate SG ($S_c$).  
The \underline{`SIMULATE'} prompt to VLM: 
\textit{``Simulate the given action sequence from the initial state, incrementally updating the scene graph."} modifying nodes, attributes, and edges of SG.
We use this initial context SG $S_C$ (same for all MCQ options) to \underline{`SIMULATE'} each MCQ option independently producing for final SG $S_m$ for the `m-th' option. 

\vspace{4pt}
\noindent \textbf{3) Similarity Comparison}: 
After simulating each action, \textit{VLM is asked to compare} \underline{`SIMILARITY'} between resultant SG $S_m$ and goal SG $S_g$, via \textit{``\textbf{Compare} the resulting scene graph with the goal scene graph to identify incorrect relationships, misplaced objects, or unmet constraints and score them between 0-100."}. 
The option that the VLMs \textit{think} is most similar ($S_m$ and $S_g$) is chosen as its prediction. 
\textit{Note:} 
SG(s) are model-specific internal representations; hence compared via model-as-a-judge, as there are no universal metrics for all models.

\subsection{Results}

\noindent \textbf{\CPP Results:}
\figno{\cref{tab:sgi_results}} shows that SGI consistently outperforms vanilla Scene Graph across all tasks.
For Step Completion, SGI yields an average gain of 4.4\%, with improvements up to 10.3\%, 4.8\%, and 10.0\% for Intern-VLM 2, Intern-VLM-3, and GPT-4o. 
For Error Detection, gains average 4.1\%, 1.7\%, and 9.2\% respectively. 
On Blocks-World-E, \textit{Gemini-2.5-pro}~\cite{team2023gemini} improves from 67\% with CoT to 70\% with SG to 71.5\% with SGI.
Although SGI requires one VLM call per step, \hl{iterating over text-based SG is far cheaper than synthesizing intermediate images}.
Total compute cost is the length of the initial context + number of MCQ options $\times$ Avg. number of steps per option.
\hl{Without stepwise updates, performance is near random for Intern-VLM 2}.
The additional compute yields up to 13\%  improvement in error detection, demonstrating its effectiveness on \CP. 
SGI has fewer cheating instances: Intern-VLM-2 (43\% vs \textit{39\%}),
Intern-VLM-3 (41\% vs \textit{35\%}),
Qwen-3 (35\% vs \textit{30\%}), and 
CoG-VLM (27\% vs \textit{23\%}).


\begin{table*}[!t]
\caption{\textbf{Scene Graph Incremental update}: SGI improves SG, same naming convention as \cref{tab:results}. 
Blocks-World-E shortened to Blocks-E, and Robo-VQA-E to Robo-E.
Other models with SGI are evaluated in the \SUPP.
}  
\centering
\renewcommand{\arraystretch}{1.06}
\setlength\tabcolsep{4pt}
\resizebox{\textwidth}{!}{
\begin{tabular}{l | cc | cc| cc | cc || cc| cc| cc }
\specialrule{1.5pt}{0pt}{0pt}
\rowcolor{mygray}
 &  \multicolumn{8}{c||}{\textbf{Step Completion} (\% $\uparrow$) } &  \multicolumn{6}{c}{\textbf{Error Detection} (\% $\uparrow$)}  \\
[-.4pt]\hhline{~--------------} 
\rowcolor{mygray}
&  
\multicolumn{2}{c|}{\textbf{Robo-E}} & 
\multicolumn{2}{c|}{\textbf{Shuffle-E}} & 
\multicolumn{2}{c|}{\textbf{Maze-E}} & 
\multicolumn{2}{c||}{\textbf{Blocks-E}} &
\multicolumn{2}{c|}{\textbf{Robo-E}} &
\multicolumn{2}{c|}{\textbf{Maze-E}} &
\multicolumn{2}{c}{\textbf{Blocks-E}}  
\\
\rowcolor{mygray} 
\multirow{-3}{*}{\textbf{Method}} & 
SG & \textbf{SGI} & 
SG & \textbf{SGI} & 
SG & \textbf{SGI} & 
SG & \textbf{SGI} & 
SG & \textbf{SGI} & 
SG & \textbf{SGI} & 
SG & \textbf{SGI} \\ 
\hline 
Random & 
\multicolumn{2}{c|}{\RC{20}}
& \multicolumn{2}{c|}{\RC{20}}
& \multicolumn{2}{c|}{\RC{20}}
& \multicolumn{2}{c||}{\RC{20}}
& \multicolumn{2}{c|}{\RC{25.4}}
& \multicolumn{2}{c|}{\RC{26.1}}
& \multicolumn{2}{c}{\RC{26.1}}\\
\hline
Intern-VLM 2 & 
25.1 & \cellcolor{green!20} 32.1& 
23.4 &\cellcolor{green!20} 25.2 &
41.2 &\cellcolor{green!20} 43.2 &  
\RC{18.9} &\cellcolor{green!20} 29.2&
\RC{26.1} & \cellcolor{green!20}31.5&  
33.4 & \cellcolor{green!20}34.8& 
37.3 & \cellcolor{green!20}42.9\\

Intern-VLM 3 & 
31.4& \cellcolor{green!20}33.7& 
27.1& \cellcolor{green!20}28.6&
50.1& \cellcolor{green!20}54.8&  
29.4& \cellcolor{green!20}30.6&
28.1& \cellcolor{green!20}29.5 &  
35.1& \cellcolor{green!20}36.3& 
44.3& \cellcolor{green!20}45.1 \\ 
GPT-4o & 
52.2 & \cellcolor{green!20} 56.4 & 
30.1 & \cellcolor{green!20}37.0  &
46.1 & \cellcolor{green!20}56.1  & 
54.3 & \cellcolor{green!20}55.3  &
44.2 & \cellcolor{green!20}57.4 & 
35.3 & \cellcolor{green!20}41.1  &
42.1 & \cellcolor{green!20}50.7 \\
\specialrule{1.5pt}{0pt}{0pt}
\end{tabular}}
\label{tab:sgi_results}
\end{table*}

\begin{table*}[!t]
\centering
\begin{minipage}[c]{0.49\textwidth} 
\renewcommand{\arraystretch}{1.05}
\caption{\textbf{Oracle and Noisy SG:} 
Robo-VQA-E has no Oracle (GT) SG.}  
\setlength\tabcolsep{4pt}
\resizebox{\textwidth}{!}{
\begin{tabular}{c| c | cc | cc| cc }
\specialrule{1.5pt}{0pt}{0pt}
\rowcolor{mygray}
& &  
\multicolumn{2}{c|}{\textbf{Shuffle-E}} & 
\multicolumn{2}{c|}{\textbf{Maze-E}} & 
\multicolumn{2}{c}{\textbf{Blocks-E}} 
\\
[-.4pt]\hhline{~~ -- -- --} 
\rowcolor{mygray} 
& \multirow{-2}{*}{\textbf{SG}} & 
SG & \textbf{SGI} & 
SG & \textbf{SGI} & 
SG & \textbf{SGI}  \\ 
\hline 
& Noisy & 20.3& 23.4&38.5 & 41.3& 15.3& 25.4 \\
& VLM & 23.4 & 25.2 & 41.2 & 43.2 & 18.9 & 29.2\\ 
\multirow{-3}{*}{\rotatebox[]{90}{Intern-2}} & GT & 23.7 & \BR{26.5} & 42.0 & \BR{59.0} & 28.1 & \BR{34.1}\\ 
\hline 
& Noisy  & 27.3 & 30.6 & 30.5 & 43.6& 24.3 & 28.7  \\
& VLM &30.3& 31.4 &36.2 & 44.6 & 28.3 & 29.5  \\ 
\multirow{-3}{*}{\rotatebox[]{90}{Qwen-3}} & GT & 30.7 & \BR{31.8} & 40.5 & \BR{55.4} & 30.3 & \BR{35.6}\\ 
\specialrule{1.5pt}{0pt}{0pt}
\end{tabular}}
\label{tab:oracle}
\end{minipage}
\hfill
\begin{minipage}[c]{0.49\textwidth} 
\renewcommand{\arraystretch}{1.4}
\caption{\textbf{SGI on VQA~\cite{wang2024pictureworththousandwords}}. 
SGI generalization on VQA tasks.}  
\setlength\tabcolsep{1pt}
\resizebox{\textwidth}{!}{
\begin{tabular}{c | ccc | ccc | ccc }
\specialrule{1.5pt}{0pt}{0pt}
\rowcolor{mygray} 
 & 
\multicolumn{3}{c|}{\textbf{Spatial-Map}} & 
\multicolumn{3}{c|}{\textbf{Maze-Nav}} & 
\multicolumn{3}{c}{\textbf{Spatial-Grid}} \\ 
[-.4pt]\hhline{~ --- --- ---} 
\rowcolor{mygray} 
\multirow{-2}{*}{\textbf{Method}} & 
CoT & SG & \textbf{SGI} &
CoT & SG & \textbf{SGI} & 
CoT & SG & \textbf{SGI} \\ 
\hline 
\makecell{CoG\\-VLM} & 
25.1 & \BR{36.7} & 35.8 & 
32.3 & \BR{32.4} & 31.2 & 
30.1 & 34.3 & \cellcolor{green!20}38.2\\
\hline
\makecell{Janus\\-pro-7B} & 
42.4 & 47.4	& \cellcolor{green!20}47.8 & 
20.8 & 27.3 & \cellcolor{green!20}29.3 & 
34.4 & 35.8 &\cellcolor{green!20} 36.3 \\
\hline
\makecell{Intern\\-VLM} & 
36.3 & 41.3	&\cellcolor{green!20} 44.3 & 
28.6 & 40.5 &\cellcolor{green!20} 42.1 & 
33.3 & 33.8 & \cellcolor{green!20}35.1 \\
\specialrule{1.5pt}{0pt}{0pt}
\end{tabular}
}
\label{tab:external_dataset}
\end{minipage} 
\end{table*}

\vspace{4pt}
\noindent
\textbf{Scene Graph Quality:}
\figno{\cref{tab:oracle}} analyzes the impact of scene graph quality on SGI. We extract ground-truth (GT, \textit{`Oracle'}) graphs for Block-World-E and Maze-E and compare them against a noisy setting, where the VLM perturbs object positions to introduce structural errors in scene graph. The results show that Oracle SG consistently outperforms VLM-generated and noisy inputs, highlighting the \insight{perception challenge: the difficulty in accurately understanding scenes}. However, even with Oracle SG, performance remains below text-only reasoning, indicating that \insight{simulating visual steps} is inherently difficult. Furthermore, the higher robustness of SGI for noisy SG in Qwen 3 demonstrates that the model is not blindly picking option A anymore.
This is because each option is independently iterated upon to derive the best option, rather than simply asking VLM to pick one.  

\begin{figure*}[!t]
\begin{minipage}[c]{0.33\textwidth} 
\centering
\captionof{table}{\textbf{ Planbench~\cite{valmeekam2023planning}}
Constrained plan generation} 
\setlength\tabcolsep{3pt}
\resizebox{\textwidth}{!}{
\begin{tabular}{c c | c }
\specialrule{1.5pt}{0pt}{0pt}
\rowcolor{mygray}
 &  & \textbf{Task 8} \\
\rowcolor{mygray} 
\multirow{-2}{*}{\textbf{Method}} & \multirow{-2}{*}{\textbf{Variant}} &  
Score ($\uparrow$) \\
\hline 
\multirow{3}{*}{\makecell{Qwen2\\ VL-8B}} 
& Vanilla & 13.8 \\
 & CoT & 14.1 \\
& SG & 13.9 \\
\rowcolor{mygray}
\textit{(our)} & SGI & 14.7 \\
\specialrule{1.5pt}{0pt}{0pt}
\end{tabular}
}
\label{tab:planbench_results}
\end{minipage}
\hfill 
\begin{minipage}[c]{0.25\textwidth} 
\centering
\captionof{table}{\textbf{Text vs MCQ}
}
\renewcommand{\arraystretch}{1.2}
\setlength\tabcolsep{2pt}
\resizebox{1.1\textwidth}{!}{
\begin{tabular}{c| c | c c}
\specialrule{1.5pt}{0pt}{0pt}

& \multicolumn{3}{c}{\cellcolor{mygray}  \textbf{Blocks-World-E}}\\ 
[-.4pt]\hhline{~---} 
& \cellcolor{mygray}  Variant & \cellcolor{mygray}  MCQ & \cellcolor{mygray}  Text  \\ 
[-.4pt]\hhline{~---} 
& CoT & 37.9 & 40.0 \\
& SG & 37.3 & 38.0 \\
\multirow{-5}{*}{\rotatebox[]{90}{Intern-VLM-2}} & \cellcolor{mygray}  SGI & \cellcolor{mygray}  42.9 & \cellcolor{mygray}  44.0 \\
\specialrule{1.5pt}{0pt}{0pt}
\end{tabular}
}
\label{tab:open_qa}
\end{minipage}
\hfill 
\begin{minipage}[c]{0.36\textwidth} 
\centering
\includegraphics[width=\linewidth]{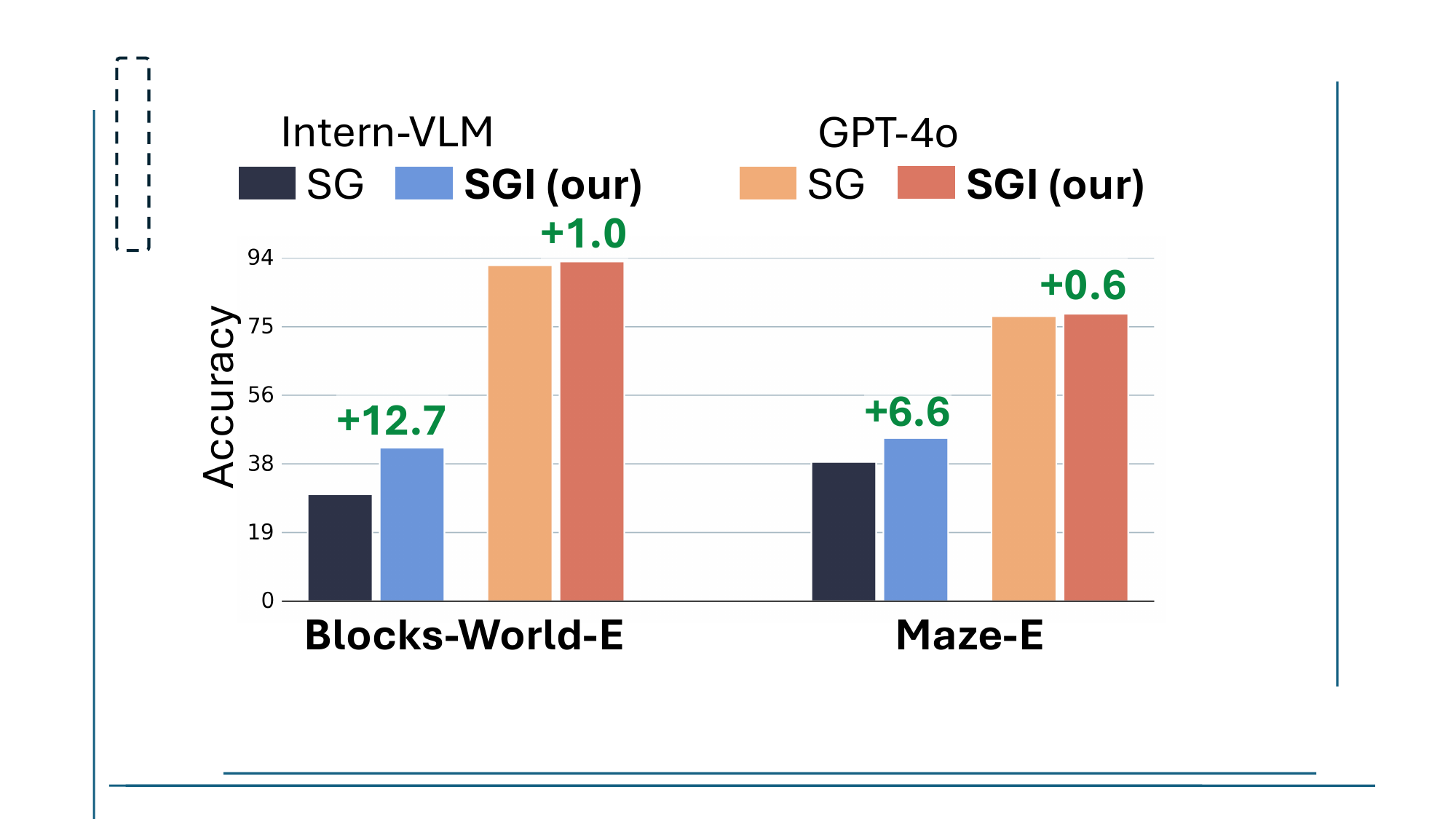}
\caption{\textbf{Error Free}}
\label{fig:proposed_error_free}
\end{minipage}
\end{figure*}

\vspace{4pt}
\noindent \textbf{External Dataset:} 
Unlike sequence planning, VQA~\cite{wang2024pictureworththousandwords} is static and lacks temporal structure. In our setup, the same image serves as both initial and final state, and the VLM simulates each MCQ option for feasibility. Iteratively evaluating options outperforms single-shot SG or CoT (\figno{\cref{tab:external_dataset}}).
We also evaluate the text-only PlanBench benchmark~\cite{valmeekam2023planning}, specifically Blocksworld Task 8, where the goal changes after the initial context, requiring the model to adapt its plan. Unlike MCQ settings, PlanBench requires generating a full action sequence that is validated by the planning environment. Qwen2-VL-8B with SGI on textual scene graphs achieves the best performance (\figno{\cref{tab:planbench_results}}). Other PlanBench tasks focus on plan generation and are out of scope. Details in \SUPP.
%

 \vspace{4pt}
\noindent \textbf{MCQ Alternative:} 
\figno{\cref{tab:open_qa}} 
shows the performance of \textbf{error detection task} in an open text generation. 
Given initial context, instead of picking an MCQ option, VLMs are prompted to write the erroneous step as a question-answering task, judged by Qwen-2 8B for similarity between ground truth and generated text.
\textbf{SGI} outperforms baselines, indicating the generalizability of the algorithm.

\vspace{4pt}
\noindent \textbf{Error-Free Scenarios:}
\figno{\cref{fig:proposed_error_free}} shows SGI outperforming SG on both error-prone and error-free 
sequence planning,
highlighting the generalizability of SGI. 

\section{Conclusion and Future Work}
\label{sec:limitations}

We introduce \CP, a benchmark for evaluating VLMs on error-prone sequential planning from images with text instructions. 
\CPP exposes critical weaknesses in VLMs: 
near-random behavior with even a single error, poor handling of in-context mistakes, and a bias toward text-based reasoning over grounded multimodal understanding. Even strong models such as \textbf{GPT-4o} and \textbf{GPT-5.1} fail to reliably use visual context to recover correct plans.
We propose \textbf{SGI} (Scene Graph Incremental Update), which refines scene representations step-by-step after each action.
SGI substantially boosts performance across error-prone, error-free, and VQA settings compared to vanilla scene graphs, highlighting its robustness
and ability to enhance corrective sequence planning in VLMs.

Our current setup uses static image pairs. Extending \CPP to video-based planning and to an interactive setting where agents execute actions and observe updated states are natural future directions. 

\section{Limitations} Our work isolates exactly one erroneous step per sequence to facilitate controlled evaluation. While this design aids in failure analysis, we acknowledge that the single error constraint is somewhat artificial. Real-world alignment requires evaluating models in more complex, multi-error settings.

\section*{Acknowledgements}
The authors thank Steven Dick (UCF High-Performance Computing) and Vibhav Vineet (Microsoft Research) for their help and
contributions to this project.
This research has benefited from the Microsoft Accelerating Foundation Models Research (AFMR)
grant program.

    
    






\bibliographystyle{splncs04}
\bibliography{main}

\title{\CP: \underline{Co}rrective \underline{S}equence \underline{Plan}ning via Scene Graph Incremental Updates\\
\textit{(Supplementary)}}

\titlerunning{\CPP}

\author{Shresth Grover$^\dagger$\and
Priyank Pathak \and
Akash Kumar\and 
Yogesh S Rawat
}
\authorrunning{Grover et al.}

\institute{
UCF Institute of Artificial Intelligence, University of Central Florida (UCF)\\[1ex]
\email{shgrover@ucsd.edu}
\hspace{2pt}
\email{\{priyank, akash.kumar, yogesh\}@ucf.edu}\\[1ex]
\textbf{Project Page}: \url{https://shroglck.github.io/cos_plan/} \\
\textbf{Dataset}: \url{https://huggingface.co/datasets/shrg7/COSPLAN} 
\blfootnote{$^\dagger$work done as intern}
}

\maketitle

\section*{Overview}
\begin{figure}[!h]
    \centering
    \includegraphics[width=\linewidth]{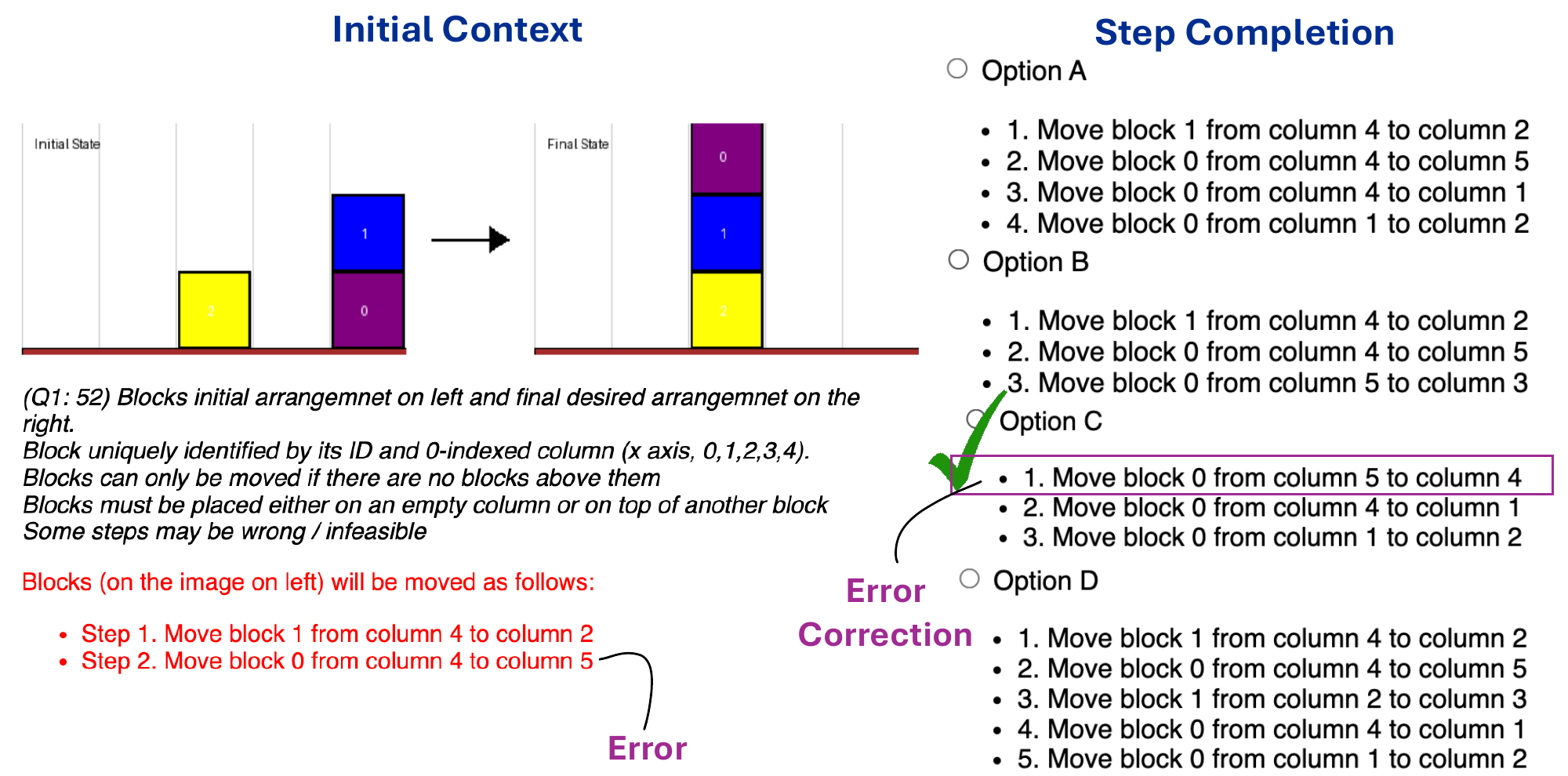}
    \caption{\textbf{\CPP overview}: 
    The input context comprises executed actions and both the initial and final states. The model predicts the optimal action steps to reach the goal and identifies errors in the provided context. The Main Submission also showed an example in Figure 1. 
    Examples like these were provided to humans as well, for calculating human scores. Above is a Block-World-E example. 
    }
    \label{fig:paper_overview3}
\end{figure}

\begin{figure}[!h]
    \centering
    \includegraphics[width=\linewidth]{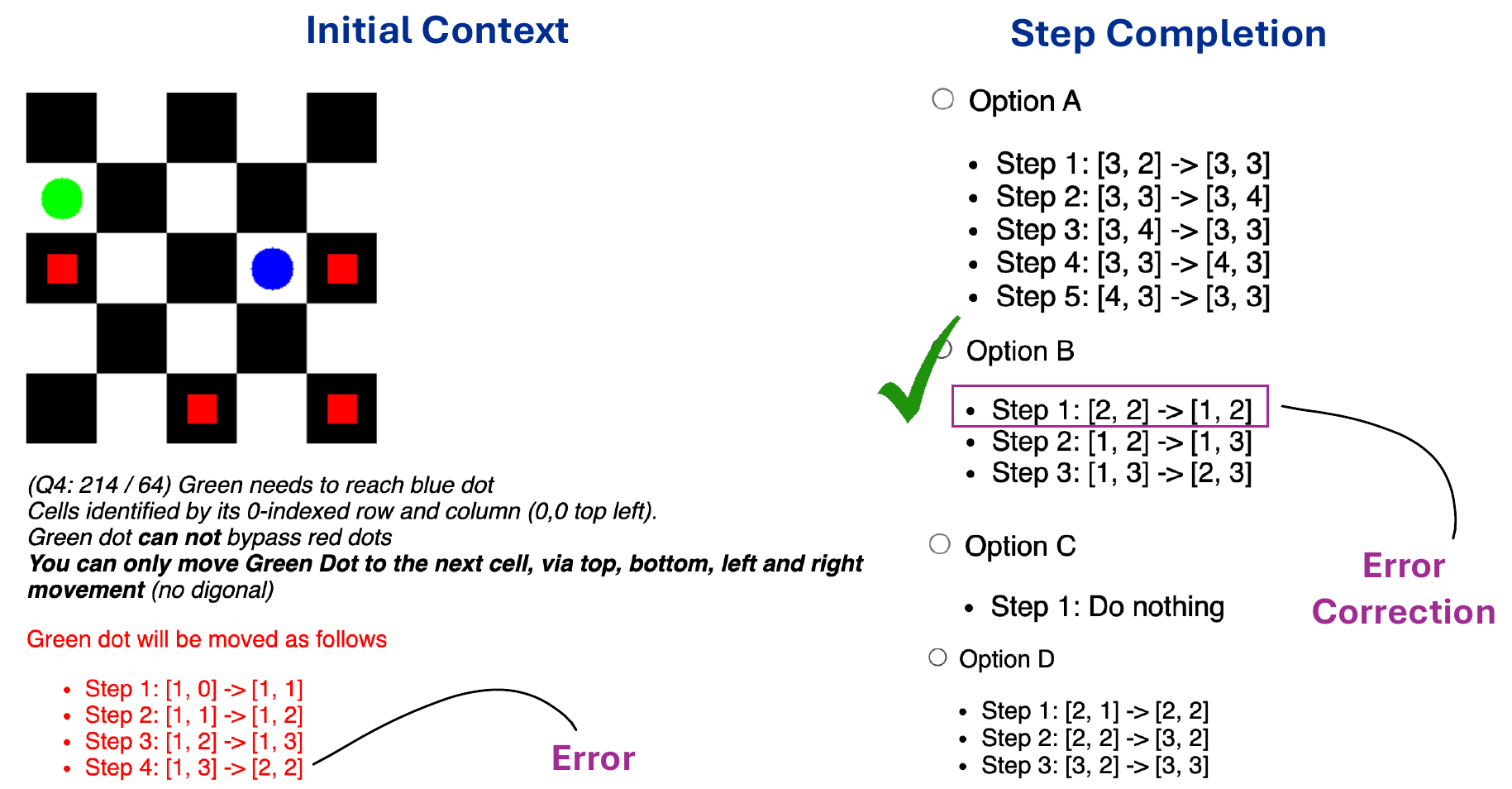}
    \caption{\textbf{\CPP overview}: 
    Another example, like \cref{fig:paper_overview3} but for MAZE-E.
    }
    \label{fig:paper_overview4}
\end{figure}

\begin{enumerate}
    \item {Limitations}
    \item \cref{fig:paper_overview3}, and \cref{fig:paper_overview4} are examples of \CP, as given to humans and models. 
    \item \cref{sec:design_choices} various design choices of \CP.
    \item \cref{sec:SGI_details} highlights more in-depth details of SGI algorithm.
    \item \cref{sec:models} has details for human baseline scores and models. 
    \item \cref{sec:exps} has results and additional experiments on \CP.
    \item 
    \cref{sec:implement} Implementation details like GPUs configuration for our benchmark.
    \item \cref{sec:ethics} lists ethical responsibility in using our \CPP benchmark. 
    \item \cref{sec:future_work} highlights the next set of experiments for the generalization of the SGI algorithm. 
    
\end{enumerate}

\section{\CPP Design Choices}
\label{sec:design_choices}

\subsection{Error Correction Design}

Current VLMs struggle with real-world deployment because their self-supervised training rarely includes the suboptimal steps or execution errors commonly encountered in autonomous navigation and robotics during inference. To accurately diagnose the performance of these models under real world conditions, we need to study their ability to reason accurately in a sequential manner. To achieve this, \CPP introduces an error in the initial context encountered by the model. This: i) allows us to mimic the real-world setting where, during execution, the model encounters suboptimal or erroneous steps; ii) the presence of errors forces the model to reason about each step. Rather than explicit instruction, the agent must autonomously recognize the need to correct the course if it deems there is a suboptimal step. The ``correct'' option recovers from the error state to the goal, while incorrect options perpetuate the failure or ignore the error entirely just aligning the steps with thee goal.

Keeping this integrated design, we evaluate \textit{Error Detection} and \textit{Step Completion} separately to distinguish between error detection failures and step completion failures. This granular analysis prevents confounding variables, such as a model's ability to spot an error versus its ability to fix it, from masking specific weaknesses (\eg, high detection but low completion scores in GPT-4o).

\subsection{Sequence Completion Design}
\CPP design choice for error correction within step completion mimics general scenarios where agents must detect and recover from errors in ongoing sequences, while completing the task. 
Alternatives like separating tasks into (i) explicit error correction and (ii) continuation from a valid state
assume the error-free steps for reaching the goal, which may not reflect practical decision-making.
Instead, our `correct' option may begin from the erroneous state but proposes a recovery sequence that leads to the goal, without additional errors (course correction). 
Similarly, `incorrect' options may perpetuate the error or introduce new ones.


\subsection{One error Design}
\CPP reveals near-random predictions of VLMs in error-prone sequential planning with \textit{just one error}. 
Expanding our analysis to multiple error cases requires automation because of the exponential complexity caused by the cascading effect of multiple errors. 
Correcting and evaluating requires automation, because of multiple (exponential) plausible corrections. 
We are currently limited by automation via intelligent VLMs (\eg GPT-4o, GPT-5.1) not being able to handle even one error.

\subsection{Scene Graph Design (SG)}
We have standardized attributes (\eg nodes for objects, edges for relations) via unified prompts, rejecting invalid formats. 
To ensure fairness across models, identical SG schemas and prompts were enforced across models, with strict JSON validation for outputs.
Cross-model comparisons thus focus on task performance under consistent structures, despite inherent differences, \eg GPT-4o vs. Qwen2 VL-8B verbosity.
Table 6 in main submission also shows Oracle SG, which is the ground truth on which synthetic problems were created, is common for all models.

\subsection{Difference between SGI vs SG \& CoT}
Chain-of-Thought (CoT) represents the most basic form, where VLMs break complex tasks into a sequence of step-by-step reasoning steps.
Scene Graph (SG) builds on CoT via a structured representation of the scene, enabling more coherent tracking and reasoning. 
Effectively, SGI interpolates CoT and SG reasoning across sequential scenes, allowing VLMs to reason through evolving scenes rather than interpolating scene-level decisions.
In terms of reasoning hierarchy, CoT $\subseteq$ SG $\subseteq$ SGI. 
Examples of CoT, SG, and SGI are attached separately.

\subsection{Chain of Though prompt Details}
 Chain-of-Thought (VisualCoT) Setup. As part of our baseline evaluations, we follow the methodology proposed in VisualCoT \cite{shao2024visualcotadvancingmultimodal}. Instead of directly predicting the final corrective action, models are prompted to generate an explicit, step-by-step intermediate reasoning chain based on the visual input. This allows the Vision-Language Models to break down the task, evaluate the current state against the goal, and identify the erroneous step before selecting the final option from the multiple-choice question.

\subsection{Annotation Generation}
To construct the \CPP benchmark, ground-truth action sequences for the synthetic environments (Maze, Blockworld, and Shuffle) were generated via Depth-First Search (DFS), ensuring optimal pathing.. To evaluate corrective sequential planning, we systematically introduce errors into these ground-truth sequences and the required correction inherently reverses or rectifies this specific invalid step.

Robo-VQA-E is manually annotated samples from Robo-VQA dataset. 
\begin{enumerate}
    \item Qwen filters videos if scene changes 
    \item Qwen identifies objects in video 
    \item Leverage the existing fine-grained annotations provided in the dataset to get correct sequence
    \item Errors is generated by randomly placing objects in templates as explained in main submission in section 3.2 Erroneous Step.

Error correction simply reverses the erroneous step. 
    
\end{enumerate}

\section{Scene Graph Incremental update (SGI) Details}
\label{sec:SGI_details}
\begin{algorithm}[!h]
{
\small
\caption{\textbf{SGI} (Error Detection)}
\label{alg:sgi_algorithm_error}
\textbf{Input}: Initial state $\mathcal{I}_0$, Goal state $\mathcal{I}_g$, Initial Context actions $\mathcal{A}_1,\mathcal{A}_2,..\MER..\mathcal{A}_{k<N}$
\begin{algorithmic}[1]
\REQUIRE VLM $\mathcal{M}$, Initial  Context as $MCQ$ m options. \\ 
\vspace{5pt}

\COM{\#\# 1) Vanilla Scene Graph}
\STATE $S_0 \gets \textsc{QUERY}\bigl[ \mathcal{M}(\mathcal{I}_0) \bigr]$ \hfill 
\COMM{// Obtain initial Scene Graph}
\STATE $S_g \gets \textsc{QUERY}\bigl[ \mathcal{M}(\mathcal{I}_g) \bigr]$ \hfill 
\COMM{// Obtain final Scene Graph}

\vspace{5pt}
\COM{\#\# 2) Incremental Scene Update ($S_0 \rightarrow S_c$)}
\STATE $S_c \gets S_0$  
\FOR{ $\mathcal{A}_i$ in [$\mathcal{A}_1,\mathcal{A}_2,..\MER..\mathcal{A}_{k<N}$] }
    \STATE $S_c \gets \textsc{SIMULATE} \bigl[ \mathcal{M}( S_c, \mathcal{A}_i) \bigr]$ \\
    \COMM{// $\mathcal{M}$ simulates i-th action $\mathcal{A}_i$ to incrementally update \\ intermediate context Scene Graph $S_c$ } 

    \STATE $sim_i \gets \textsc{SIMILARITY} 
\bigl[ \mathcal{M}(S_c, S_g) \bigr]$ \\
\COMM{// we compute similarity of context action $S_c$ with goal $S_g$} 
\ENDFOR

\vspace{5pt}
\COM{\#\#) Error Detection Similarity}
\STATE $sim_{i'} \gets \arg\min_{i \in S_i} [sim_i]$ \\ 
\COMM{Find the least similar context action, measures deviation. 
}
\IF{ $sim_{i'} > 0.75$}
    \STATE \textbf{return} ``None of the above"\\
    \COMM{// if the least similarity $sim_{i'}>$ 0.75 (hyperparameter), deviation from goal is not large enough for error. 
}
\ELSE
    \STATE \textbf{return} $A_{i'}$\\
    \COMM{// $A_{i'}$ produces scene graph with similarity $<0.75$ and has maximum deviation.}
\ENDIF

\end{algorithmic}
}
\end{algorithm}

The Scene Graph Incremental update (SGI) framework enhances the decision-making of VLMs in sequential instruction-following, particularly when handling incomplete plans or embedded errors (\Er). Unlike conventional Chain-of-Thought (CoT) approaches that infer the transformation from $\mathcal{I}_0$ to $\mathcal{I}_g$ in a single step, SGI decomposes reasoning into structured, interpretable updates. While the main paper details SGI for Step Completion, here we present the adaptation for Error Detection in Algorithm~\ref{alg:sgi_algorithm_error}.

Formally, given the initial state $\mathcal{I}_0$ and goal state $\mathcal{I}_g$, we derive structured scene graphs $S_0$ and $S_g$, capturing entities and spatial relations. The VLM processes a context sequence $\mathcal{A}_{1 \dots \EER \dots k<N}$ containing an error \Er. SGI operates via 
\textit{Incremental Simulation} and \textit{Similarity-Based Selection}. 

\vspace{6pt}
\noindent \textbf{Incremental Scene Update.}
We task the VLM to \underline{`SIMULATE'} the textual actions, acting as a state tracker to modify nodes and relational edges.
The simulation prompt is: \textit{``Simulate the given action sequence from the initial state, incrementally updating the scene graph.``}
The model propagates the context actions $(\mathcal{A}_1, \dots, \MER, \dots, \mathcal{A}_k)$ over the initial graph $S_0$ to generate an intermediate graph $S_c$, representing the environment state after the context actions.
Subsequently, for each available MCQ candidate option \textit{`m'}, the model applies the proposed action to $S_c$, generating a hypothetical resultant graph $S_{m}$.
Since VLMs simulate all the steps, the simulation depth is fixed for all the models. Additionally, we do not supervise SG generation, \ie nodes, and edges can sometimes be noisy.

\vspace{6pt}
\noindent \textbf{Similarity Comparison.}
To identify the optimal continuation, SGI compares the hypothetical graph $S_{m}$ against the ground truth goal graph $S_g$.
The VLM is prompted: \textit{``Compare the resulting scene graph with the goal scene graph to identify incorrect relationships, misplaced objects, or unmet constraints. Select the best-aligned plan."}
The system selects the option that maximizes structural alignment with $S_g$, ensuring decisions are grounded in verifiable physical constraints rather than superficial text probabilities.
The similarity scores are based on VLM judgment, which is not guided by us, since Scene-graphs are model-dependent and one universal similarity metric may not be applicable on all kind of scene graphs.

\section{Contenders for solving \CPP}
\label{sec:models}
We employ a suite of state-of-the-art vision-language models (VLMs) to address visual reasoning tasks, including both proprietary and open-source solutions. 
These models exhibit diverse architectural characteristics for multimodal understanding.

\subsection{LLaMA-3} LLaMA-3 \cite{llama}
is a general-purpose VLM that uses textual and visual inputs. It integrates both modalities within a single architecture for text and image understanding. 

\subsection{GPT-4o \& GPT-5.1}
GPT-4o \& GPT-5.1~\cite{gpt4} is a general-purpose VLM that uses textual inputs and outputs. It    integrates modality processing within a single model architecture for text, images, and audio processing.


\subsection{CoG-VLM}
CoG-VLM~\cite{cogvlm} (Cognitive Vision-Language Model) is designed for visual reasoning tasks with enhanced spatial understanding capabilities.\textbf{Vision Backbone}: Utilizes EVA2-CLIP-E as the ViT encoder with the final aggregation layer removed to preserve spatial information.\textbf{Language Model}: Built on Vicuna1.5-7B, with causal masking for attention operations.

\subsection{InternVL2-26B}
InternVL2-26B~\cite{internvl} is a multimodal model optimized for visual understanding and reasoning tasks.\textbf{Architecture}: Combines InternViT-300M-448px for vision processing with internlm2\_5-7b-chat for language tasks.

\subsection{Qwen2-VL}
Qwen2-VL~\cite{qwen} is a multimodal model from the Qwen family, utilizing a 7B parameter variant.\textbf{Language Backbone}: Based on Qwen2-7B large language models.

\subsection{Janus}
Janus is a vision-language model designed for understanding, generation tasks, and specially multi modal reasoning represented by the Janus-Pro variant.\textbf{Vision Understanding}: Employs SigLIP as a vision encoder for semantic feature extraction. \textbf{Language Processing}: Implements a transformer-based language model.

\subsection{Human baseline}
Humans were shown examples like \cref{fig:paper_overview3} \& \cref{fig:paper_overview4} with a total of 58 submissions, with each contestant solving a subset of puzzles.
The stats for computing the human baseline 
are shown in \cref{tab:human_base}.

\begin{table*}[!h]
\centering
\caption{\textbf{Human Baseline Stats} Unique means unique number of puzzles solved, since some participants solved overlapping repeated submissions. 
}
\setlength\tabcolsep{4.9pt}
\scalebox{1}{
\begin{tabular}{l cccc}
\specialrule{1.5pt}{0pt}{0pt} 
\rowcolor{mygray}   
\textbf{Dataset}  & 
\textbf{Unique Puzzles} & \textbf{Correct} & \textbf{Total} & \textbf{Accuracy (\%)}  \\ 
\hline 
Shuffle-E & 45 & 36 & 67 & 53.7 \\ 
Robo-VQA-E &  17 & 8 & 19 & 42.1\\ 
Maze-E  & 22 & 22 & 23 & 95.7 \\ 
Blocks-World-E & 19 & 18 & 22 & 81.8 \\

\specialrule{1.5pt}{0pt}{0pt}
\end{tabular}
}
\label{tab:human_base}
\end{table*}

\subsection{Experimental Setup}
\begin{itemize}
    \item \textbf{Input Preparation}: Raw images and text prompts are converted into model-specific input formats, with concatenated initial and target images.
    \item \textbf{Query Formation}: Structured as \texttt{\{information about the env\} -> Task} to optimize reasoning capabilities.
    \item \textbf{Output Processing}: Model responses are parsed into structured formats for evaluation metrics.
\end{itemize}

\begin{table*}[t]
\caption{\textbf{All performance on \CPP benchmark:} Same naming convention as that of Table 3 from main submission. \textbf{SGI} results same as Table 5. 
Robo-VQA-E represented as Robo-E, and Blocks-World-E as Blocks-E.
$\dag$ Since GPT-5.1 has a similar performance as that of GPT-4o, we have skipped Error detection because of budget constrainsts.
}

\centering
\renewcommand{\arraystretch}{1.06}
\setlength\tabcolsep{2pt}
\scalebox{0.86}{
\begin{tabular}{l c | cc cc | cc c}
\specialrule{1.5pt}{0pt}{0pt}
\rowcolor{mygray}
& &\multicolumn{4}{c|}{\textbf{Step Completion} (\% $\uparrow$)} &  \multicolumn{3}{c}{\textbf{Error Detection} (\% $\uparrow$)}  \\
[-.4pt]\hhline{~~-------}
\rowcolor{mygray}
\rowcolor{mygray}
\multirow{-2}{*}{ \textbf{VLM} 
} &
\multirow{-2}{*}{\textbf{Method}} & 
\textbf{Robo-E} &
\textbf{Shuffle-E} &
\textbf{Maze-E} &
\textbf{Blocks-E} &
\textbf{Robo-E} &
\textbf{Maze-E} &
\textbf{Blocks-E}
\\ 
\hline
\hline
Random & & 20 & 20 & 20  & 20 & 15.4 & 33.3 & 33.3 \\ 
\hline
\multirow{4}{*}{Qwen2 VL-8B } & Vanilla & 17.1 & 24.1 & 26.5 & 18.1 & 9.2  & 20.5 & 32.3 \\ 
& CoT & 17.6 & 24.9 & 27.9 & 18.6 & 9.1 & 
20.8 & 30.6\\
& SG & 18.9 & 25.1 & 28.3 & 18.8 & 9.6 & 20.7 & 35.2 \\
& SGI & 19.1 & 25.0 &  28.5 & 18.5 & 10.1 & 21.3 & 35.7\\
\hline 
\multirow{4}{*}{CoG-VLM} & Vanilla & 13.1 & 23.1 & 25.1 & 25.5 & 32.1 & 6.4 & 41.3 \\ 
& CoT & 12.5 & \HB{27.1} & 25.9 & 25.2 & 33.4 & 8.4 & 43.1\\
& SG & \HB{21.5} & 23.7 & 26.5 & 26.7 & 35.3 & \HB{13.3} & 44.5 \\
\rowcolor{mygray}
\textit{(Our)} &\textbf{SGI} & 22.1 & 26.9 & 29.3 & 26.4  & 38.7 & 11.0 & 46.1 \\
\hline 
\multirow{4}{*}{Janus-pro-7B} & Vanilla & 14.1 & 23.2 & 20.4 & 24.2 & 17.5 & 20.5 & 29.3 \\ 
& CoT & 14.7 & 23.1 & 20.2 & 23.1 & 18.1 & 19.1 & 31.0 \\
& SG & \HB{21.3} & 23.5 & 21.7 & 25.1 & \HB{26.1} & 21.0 & \HB{27.6} \\
\rowcolor{mygray}
\textit{(Our)} & \textbf{SGI} & 21.1 & 26.1 & 23.2 & 26.3 & 27.6 & 21.6 & 33.2 \\
\hline 
\multirow{4}{*}{Intern-VLM-2} & Vanilla & 22.1 & 20.1 & 21.6 & 18.3 & 24.3 & 32.8 & 36.5 \\ 
 & CoT &23.5 & 23.2 & \HB{35.8} & 21.2 &25.2  & 33.1 & 37.9 \\
 & SG & 25.1 &  23.4 & \HB{41.2} & 18.9 & 26.1 & 33.4 & 37.3 \\
 \rowcolor{mygray}
\textit{(Our)} & \textbf{SGI} & 32.1 &  25.2  &  43.2 & 29.2 & 31.5 & 34.8 & 42.9 \\
\hline 
\multirow{2}{*}{GPT-4o}   & CoT & 48.2 & 27.6  & 45.6 & 49.7 & 45.3 & {40.3} & {35.1}\\
& SG & \HB{52.2} & 30.1 & 46.1 & \HB{54.3} & 44.2  & \HB{35.3} & \HB{42.1} \\
\rowcolor{mygray}
\textit{(Our)} & \textbf{SGI} & 56.4 & 37.0  &  56.1 & 55.3 & 57.4 & 41.1 & 50.7 \\
\hline 

\multirow{2}{*}{GPT-5.1} $\dag$ & CoT & 49.3 & 31.5  & 49.1 & 48.1 & - & - & - \\
& SG & 51.3 & 33.4 & 48.3 & 52.3 & -  & - & - \\
\rowcolor{mygray}
\textit{(Our)} & \textbf{SGI} & 57.5 & 35.7  &  55.7 & 55.6 & - & - & - \\

\specialrule{1.5pt}{0pt}{0pt}
\end{tabular}
}
\label{tab:supp_results}
\end{table*}

\section{Results}
\label{sec:exps}

\subsection{External Dataset Description}
\vspace{3pt}
\noindent \textbf{VQA External Dataset Details}
Wang \etal\cite{spatial2024eval} proposed a series of visual question-answering tasks to test VLM's ability on visual reasoning (different from our sequence planning tasks, as they involves no intial context).
\textbf{i) Spatial-Map:} Tests spatial relationships between objects with unique location names (\eg ``Unicorn Umbrellas"). 
Objects have pairwise relationships like "A is Southeast of B." Questions ask about spatial relationships and counting objects meeting spatial criteria.
\textbf{ii) Maze-Nav:} Evaluates navigation through mazes from the starting point (S) to the exit (E). Uses colored blocks (green=start, red=exit, black=walls, white=paths, blue=solution path) or ASCII representation. Questions count turns and determine spatial relationships between S and E.
\textbf{iii) Spatial-Grid:} Tests spatial reasoning in structured $5\times5$ grids containing animals (cat, dog, elephant, giraffe, rabbit). Questions involve counting specific animals and identifying animals at specific grid coordinates.
These datasets collectively focus on evaluating and advancing spatial reasoning capabilities and are publicly available. 

\vspace{6pt}
\noindent \textbf{PlanBench (Algorithmic Generalization)} 
In this evaluation setting, we test the model's ability to perform inductive reasoning over sequential actions. We define a \textit{planning problem} as a tuple consisting of an initial state and a goal configuration, and a \textit{plan} as the sequence of actions required to transition from the start to the goal. 
Unlike standard instruction following, the prompt here consists of few-shot example traces generated by a fixed underlying program, \eg a script containing latent control flows such as loops or conditionals (\eg an algorithm to ``unstack all blocks''). 
The model is tasked with generating a plan for a \textit{new problem instance} that follows this same structural logic but differs in complexity (\eg a larger number of objects requiring more iterations). 
We evaluate performance by verifying if the generated sequence is valid: it must be executable within the domain constraints and successfully satisfy the specified goal conditions.

\noindent \underline{\textbf{SGI for planning}}
\noindent
\cite{valmeekam2023planning} uses an off the shelf planner to evaluate LLM plans for known tasks like blockworld, indicating that these planners are able to achieve very high accuracy on these tasks. In our experiments, the text only version also shows almost perfect performance especially for GPT-4o indicating this is a with how VLM anticipates the sequence of actions.

\subsection{Experiments}

\noindent \textbf{Baseline}
CoT step-by-step prompting have been shown to significantly boost performance on arithmetic and logical tasks~\cite{wei2022chain}. This makes it our preferred baseline approach. SG, on the other hand, captures objects, attributes, and relationships, providing structured representations that enhance a VLM’s ability to reason about complex scenes.

\vspace{6pt}
\noindent \textbf{Table 2 results}
Comparing different VLMs, from Table 2, we observe that InternVLM and GPT-4o methods consistently outperform prior baselines on both \textit{Step Completion} and \textit{Error Detection}. 
GPT-4o achieves the highest performance across all tasks, indicating the synergy between strong underlying language and reasoning about sequence of actions as expected. 
Intern-VLM also demonstrates significant gains over its counterparts showing strong performance on Maze-E and Robo-VQA as compared to other models. 
\Cref{tab:supp_results} presents detailed results for the models providing a comperehensive coverage of models and zero shot adapting techniques.

\begin{figure}[!h]
\centering
\includegraphics[width=0.3\linewidth]{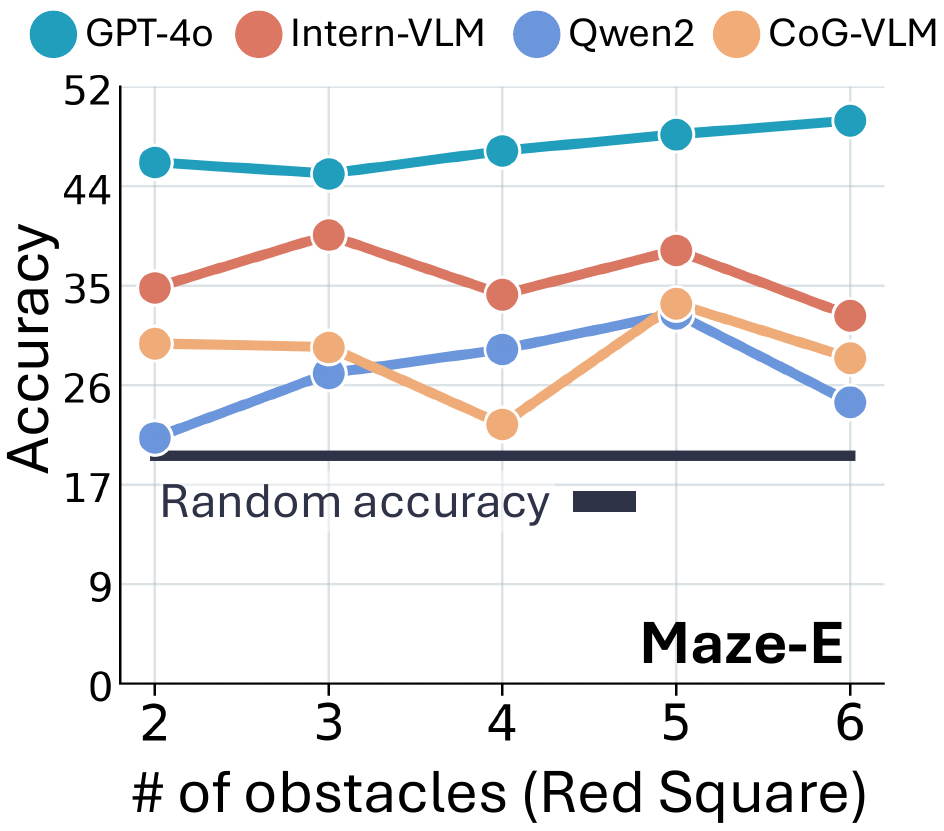}
\caption{
As the number of obstacles increases, the accuracy remains pretty much constant for all VLMs. CoT technique.}
\label{fig:no_obstacles_maze}
\vspace{-4pt}
\end{figure}

\noindent \textbf{Effect on number of Obstacles} 
\underline{\Cref{fig:no_obstacles_maze}} shows that the number of obstacles (red box) in Maze-E doesn't seem to impact performance.
This is likely because the model are not able to understanding the problem and may even ignoring the boxes/obstacles. 

\begin{figure}[!h]
\centering
\includegraphics[width=0.7\linewidth]{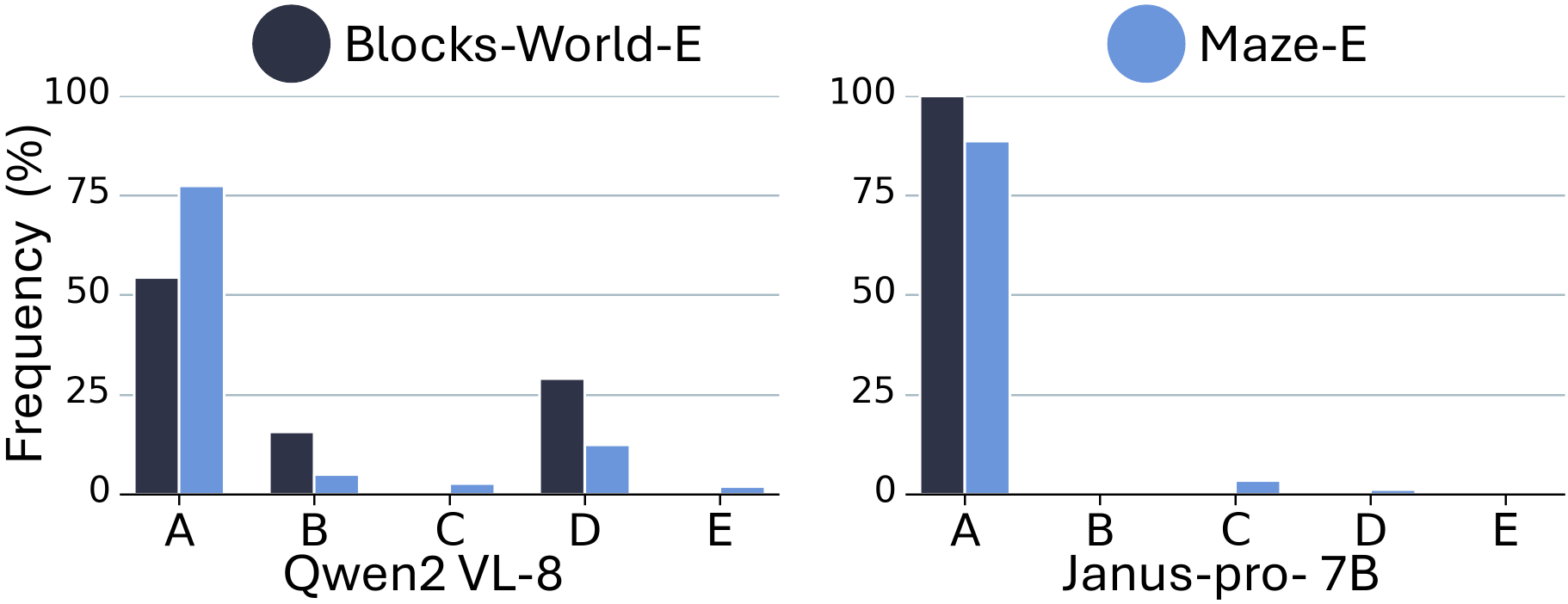}
\caption{
Models have a strong bias towards picking option A regardless of the goal and context, partially explaining the reason for random accuracy prediction.
CoT technique.
}
\label{fig:option_a_pick}
\end{figure}

\vspace{6pt}
\noindent \textbf{Bias towards selective options} 
For Blocks-World-E, Janus predicts 100\% of the time option A, and a strong bias towards the prediction of option A (\underline{\cref{fig:option_a_pick}}). This partially explains the accuracy near random, as the correct solution appears at option A with a uniform probability among all 5 options (A, B, C, D, and E).  
Similar problem with Qwen 2, where the model predicts option A irrespective of the content of option. 

\begin{figure}[!ht]
\centering
\begin{subfigure}[t]{0.47\textwidth}
\centering
\includegraphics[width=0.7\linewidth]{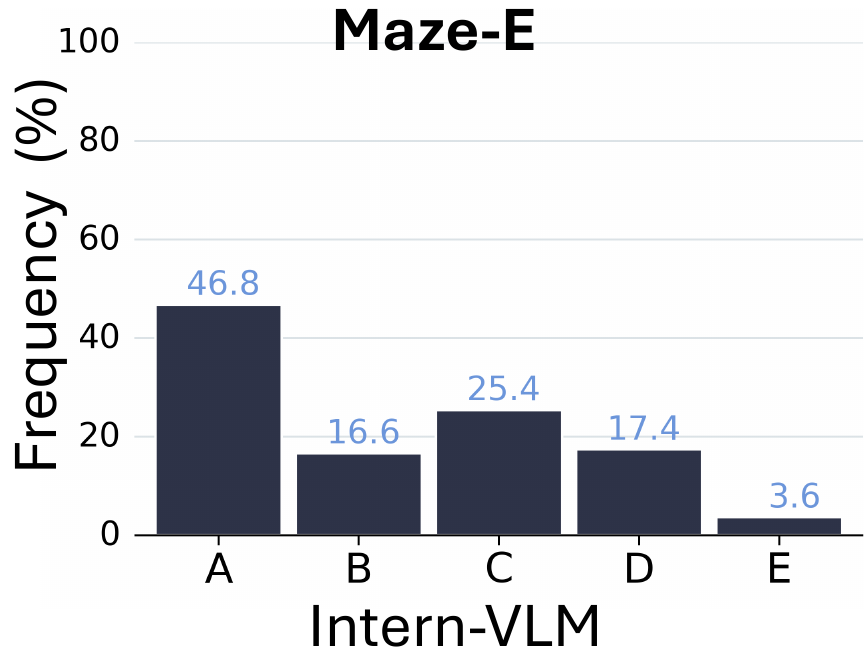}
\caption{
No of times Intern-VLM (CoT) cheat (pick the option describing final state without error correction.
}
\label{fig:chear_intervlm}
\end{subfigure}
\hfill
\begin{subfigure}[t]{0.47\textwidth}
\centering
\includegraphics[width=\linewidth]{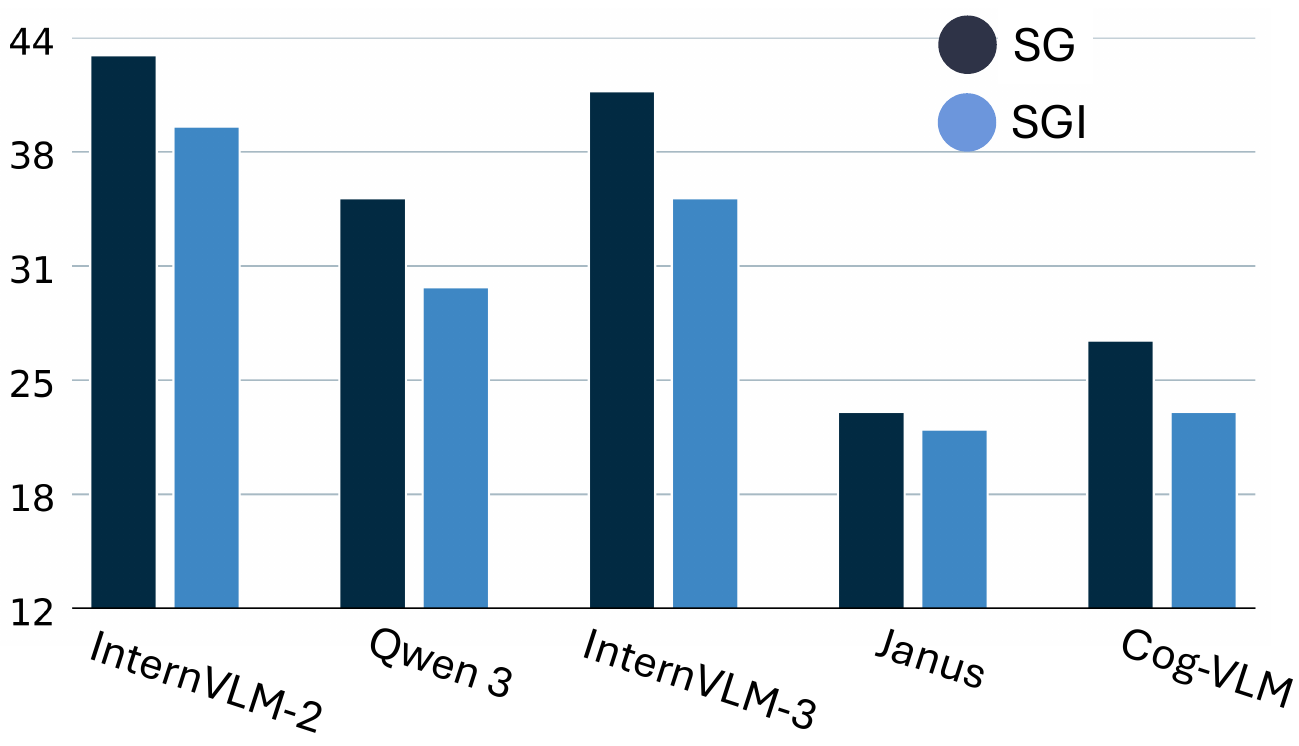}
\caption{
SGI improvement in reducing cheating scenarios across models. Maze-E dataset used. 
}
\label{fig:models_cheat}
\end{subfigure}
\end{figure}

\vspace{6pt}
\noindent \textbf{Bias towards Cheating the answers} 
\underline{\Cref{fig:chear_intervlm}} shows where Intern-VLM cheats (pick the option describing the final state, without error correction), that too with a bias towards picking option A. 
The random probability of picking an option is 20\%, which implies it's a conscious decision by the model to pick a certain option, whether a bias towards a certain option or cheating. 

\underline{\Cref{fig:models_cheat}} shows the improvement in model prediction where the SGI variant cheats less often than the simple Scene Graph (SG).
Lower cheating performance of Janus can be explained by blindly picking option A, hence the possibility of cheating is whenever cheating option appears at option A (close to random 20\% chance).

\vspace{6pt}
\noindent \textbf{Ignoring additional context:}
\begin{figure}[!h]
\centering
\includegraphics[width=0.8\linewidth]{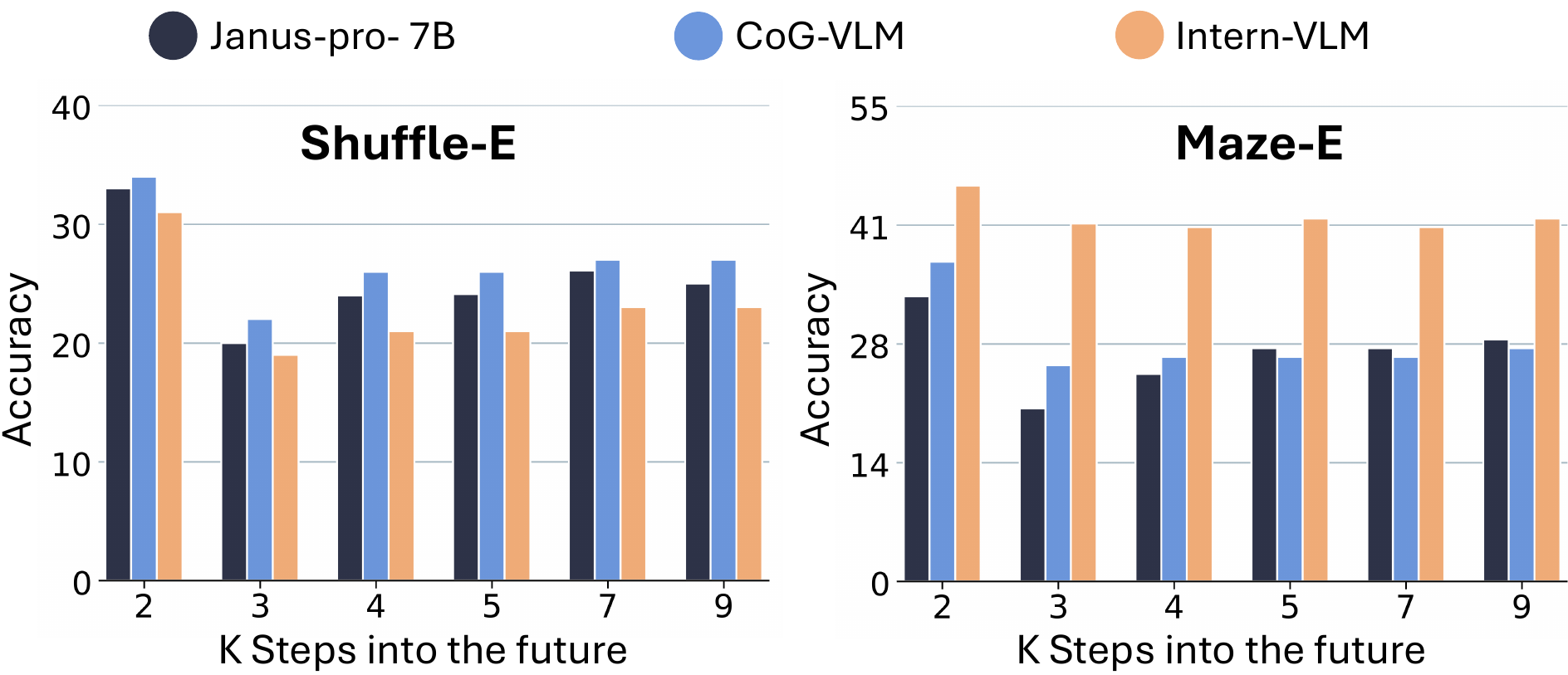}
\caption{
Accuracy of taking K steps towards the goal $\mathcal{I}_g$.
}
\label{fig:k_steps_into_future}
\end{figure}
\underline{\Cref{fig:k_steps_into_future}} generalizes the observation of how models ignores additional context  (Fig 6 (mid) in main submission). 
The setup remains the same, as the main submission, where for a constant context of length 2 (1 initial step and 1 error), performance was evaluated for step completion, where the models need to take 9 steps to reach the goal $\mathcal{I}_g$ (inclusive of 1 error correction).
Main submission showed the constant accuracy for Blocks-World-E, while here we show it for Shuffle-E and Maze-E. 
The observations remains consistent here as well, \ie models don't seem to be using the additional available context ($\propto k$) in MCQ options to reach goal $\mathcal{I}_g$. 
All models \textbf{maintain stable accuracy regardless of k},

\vspace{6pt}
\noindent \textbf{MCQ Options:}
\underline{\Cref{fig:MCQ_mze_variation}} 
generalizes the observation of Fig 6 (left)) in the main submission for Janus-pro- 7B, and CoG-VLM, showing that increasing the number of MCQ options drops the performance. 
\begin{figure}[!ht]
\centering
\includegraphics[width=0.4\linewidth]{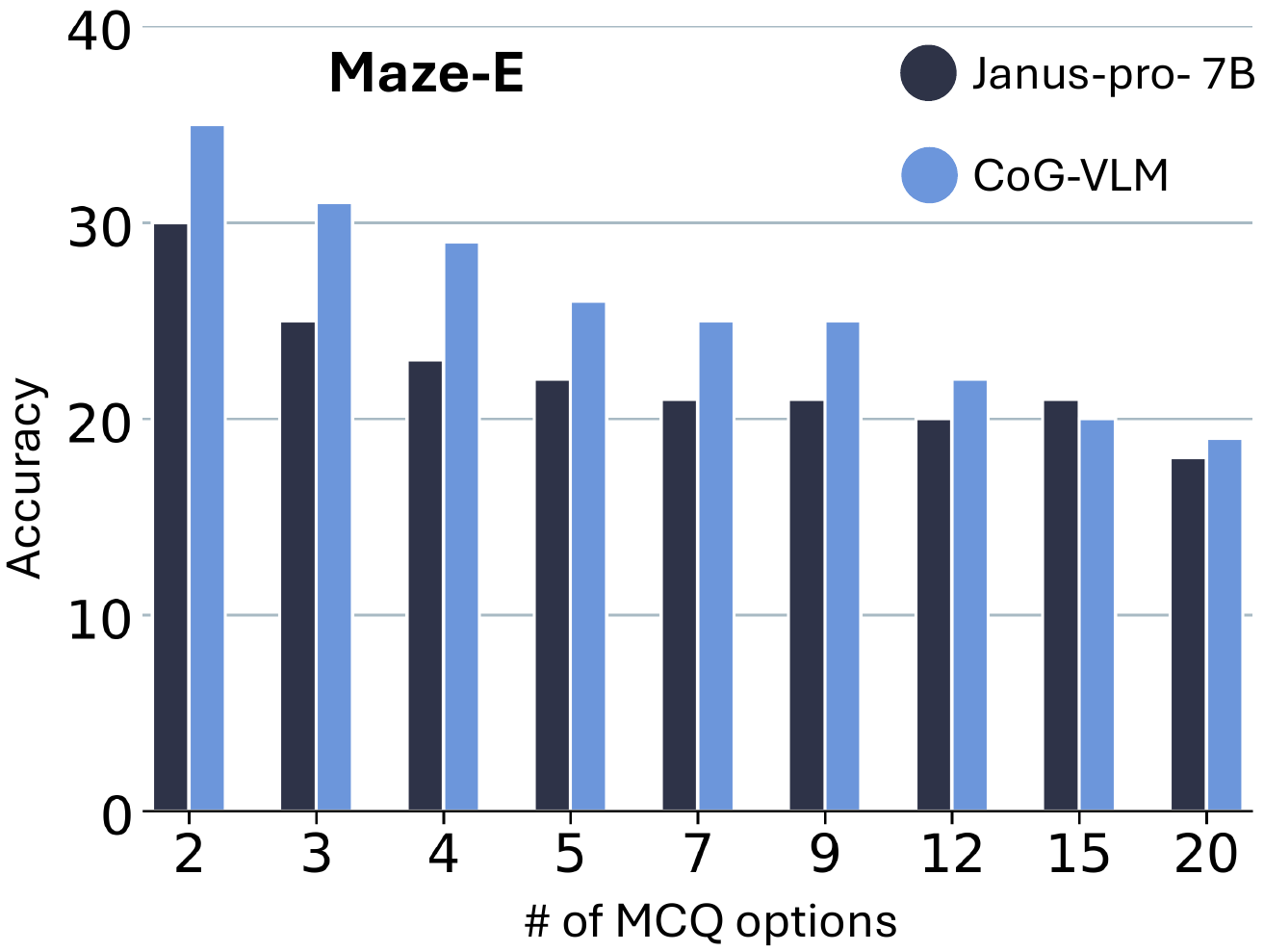}
\caption{
Same convention as Fig 6 (left) in the main submission. 
Model accuracy goes down with the number of MCQ options.
}
\label{fig:MCQ_mze_variation}
\end{figure}

\subsection{Multi Error Analysis}
While \CPP focuses on single errors for controlled analysis, we also evaluated a multi-error setting. Since multiple errors introduce exponential complexity (multiple valid correction paths), we designed a tractable setup with a fixed correction order (e.g., correct error 1, then error 2). As shown in Table \ref{tab:multi_error}, performance monotonically degrades as the number of errors increases in Maze-E, confirming that tracking compounding errors remains a significant challenge for current VLMs.

\begin{table}[h]
\centering
\small
\caption{Performance on Maze-E with increasing errors (fixed correction order).}
\label{tab:multi_error}
\begin{tabular}{lcccccc}
\toprule
\textbf{\# Errors} & \textbf{1} & \textbf{2} & \textbf{3} & \textbf{4} & \textbf{5} & \textbf{7} \\
\midrule
InternVLM (SG) & 41.2 & 40.3 & 37.5 & 33.4 & 27.4 & 22.5 \\
InternVLM3 (SG) & 50.1 & 49.2 & 46.3 & 41.5 & 36.6 & 31.3 \\
\bottomrule
\end{tabular}
\end{table}

\section{Implementation Details}
\label{sec:implement}
We set a threshold of 0.75 for similarity in error detection (\cref{alg:sgi_algorithm_error}).
The batch size was set 1.
Number of GPUs used was 1, 48Gb on a NVIDIA RTX A6000 GPU.
We would additionally release our code base for task generation and evaluation, along with our SGI algorithm.


\section{Ethical Statement}
\label{sec:ethics}

The \CPP benchmark includes both synthetic and real-world task settings. All real-world images are either synthetic or sourced under permissible licenses without depicting identifiable individuals or private information. 
While \CPP highlights the limitations of VLMs in sequential reasoning, it is not intended for deployment in safety-critical applications. Additionally, models evaluated may exhibit biases inherited from pretraining data. The dataset and code will be released for research purposes only, and we advise responsible use.

\section{Future work}
\label{sec:future_work}
All VLMs struggle with \textbf{visual + text} based sequence planning tasks, further complicated by the addition of \textbf{just one basic error}.
Optimizing the SGI algorithm for skipping certain states would be the next step in development in this iterative step-by-step decision making. 
Since our algorithm is based on the idea of simulating each step/action in a sequence, in its core, it's not really dependent on the scene graph. Future work will look into the extension of step-by-step simulation to other forms of reasoning algorithms. 
We will also be generalizing the SGI algorithm beyond static images to video-based reasoning. 

The MCQ design helps us do controlled analysis, where VLMs perform near randomly, even under the simplified MCQ setting. 
Open-ended generation is a much harder task and left as future work.

\end{document}